\documentclass[10pt,twocolumn,letterpaper]{article}

\usepackage{cvpr}              % To produce the CAMERA-READY version
\usepackage[dvipsnames]{xcolor}

\definecolor{cvprblue}{rgb}{0.21,0.49,0.74}
\usepackage[pagebackref,breaklinks,colorlinks,citecolor=cvprblue]{hyperref}
\usepackage{mathtools}
\usepackage{amsfonts}
\usepackage{tikz}
\usepackage{svg}
\usepackage[standardsections]{scrhack}
\usetikzlibrary{positioning}
\usepackage{pgfkeys}
\usepackage[raggedright]{titlesec}
\usepackage{multirow} 
\title{Dress-Me-Up: A Dataset \& Method for Self-Supervised 3D Garment Retargeting}

\newcommand*{\affaddr}[1]{#1} % No op here. Customize it for different styles.
\newcommand*{\affmark}[1][*]{\textsuperscript{#1}}

\author{%
Shanthika Naik\affmark[1], \space Kunwar Singh\affmark[1], \space Astitva Srivastava\affmark[1], \space Dhawal Sirikonda\affmark[1], \space Amit Raj\affmark[2],\\ Varun Jampani\affmark[2] , \space Avinash Sharma\affmark[1] \\
\affaddr{\affmark[1] \space International Institute of Information Technology, Hyderabad}\\
\affaddr{\affmark[2] \space Google Research}\\
}

\begin{document}
    \twocolumn[{
    \renewcommand\twocolumn[1][]{#1}
    \maketitle
    \begin{center}
    \includegraphics[width=1.0\linewidth]{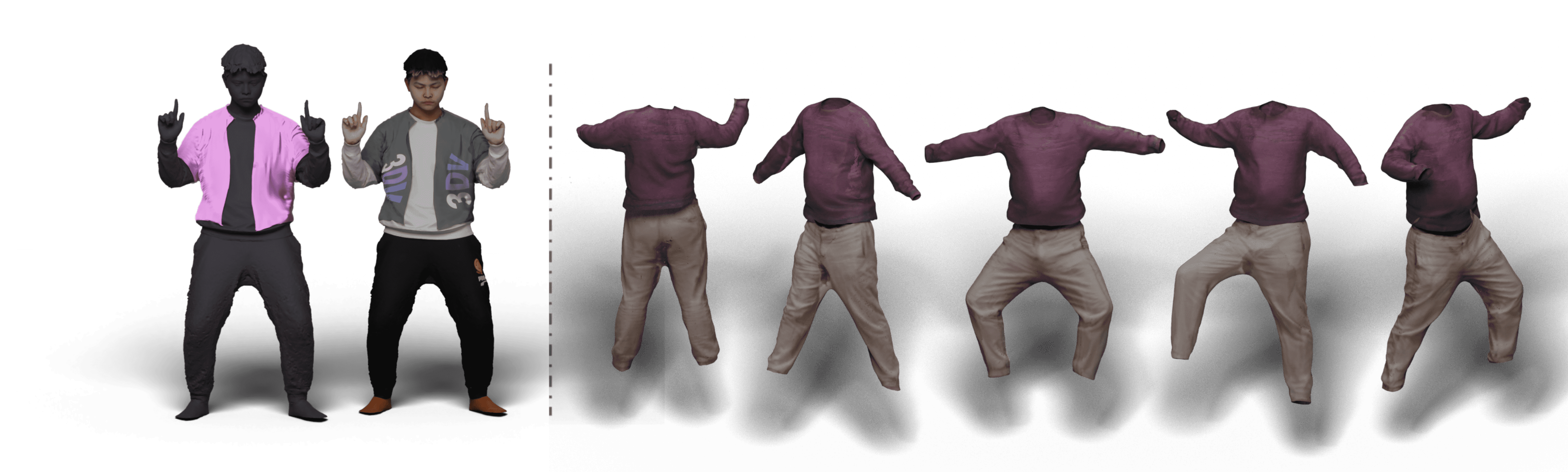}
    \captionof{figure}{
    We present Dress-Me-Up, the first-ever benchmark and dataset for retargeting non-parametric real 3D garments. As shown on left, our method can retarget arbitrary 3D garments on a non-parametric human body. On the right, we showcase a sample from our proposed real-world 3D VTON dataset. }
    \label{fig:teaser}
\end{center}
    }]
    
\begin{abstract}{
We propose a novel self-supervised framework for \textit{retargeting} non-parameterized 3D garments onto 3D human avatars of arbitrary shapes and poses, enabling 3D virtual try-on (VTON). 
% Though various self-supervised garment simulation methods exist, they don't address the problem of 3D garment \textit{retargeting}. 
% Unlike simulation, where the task is to estimate garment dynamics over the given animated sequence of poses, \textit{retargeting} requires draping an arbitrary garment to a human body in an arbitrary shape and pose. 
Existing self-supervised 3D retargeting methods only support parametric and canonical garments, which can only be draped over parametric body, e.g. SMPL. To facilitate the non-parametric garments and body, we propose a novel method that introduces \textit{Isomap Embedding} based correspondences matching between the garment and the human body to get a coarse alignment between the two meshes. We perform neural refinement of the coarse alignment in a self-supervised setting. Further, we leverage a Laplacian detail integration method for preserving the inherent details of the input garment. For evaluating our 3D non-parametric garment retargeting framework, we propose a dataset of $255$ real-world garments with realistic noise and topological deformations. The dataset contains $44$ unique garments worn by $15$ different subjects in $5$ distinctive poses, captured using a multi-view RGBD capture setup. We show superior retargeting quality on non-parametric garments and human avatars over existing state-of-the-art methods, acting as the first-ever baseline on the proposed dataset for non-parametric 3D garment retargeting.
}\end{abstract}    
    \section{Introduction}
{
    \label{sec:intro}
     3D garment modelling for virtual try-on is an active area of research with wide range of applications in fashion e-commerce and AR/VR.  A majority of deep learning methods assume the availability of synthetic parametric garment meshes~\cite{bcnet, bhatnagar2019mgn, smplicit, mir20pix2surf, majithia2022robust, deep_parametric_singleview}, while some of the nascent efforts on garment digitization~\cite{reef, xcloth} are capable of extracting high-fidelity non-parametric 3D garments from monocular images. For enabling 3D virtual try-on, the current key challenge is to perform automated retargeting of the 3D garments over digital human avatars.
     % Majority of existing vast 
     % Most existing literature on VTON focuses on 2D draping solutions \cite{HD_VTON, HR_VTON, posestyle, vton2d1, vton2d2, vton2d3, vton2d4} that operate in image space to generate reposed garment images or retarget clothing images on top of human images. However, 2D space is inherently restrictive when it comes to modelling geometric deformations or changes in viewpoints \& perspectives. 
     % Unlike image-based 2D VTON solutions\cite{HD_VTON, HR_VTON}, 3D garment retargeting offers a more controllable and elegant solution leading to an immersive experience for AR/VR environments.
     % Efforts[CITE BODYMAP] have been made in image-space which tackling Virtual try-on in the pixel space by altering texture space. While the works produce decent results, 2D space inherently is restrictive unlike 3D and do not model geometric deformations. On the other hand usage of 3D space for the tackling the Virtual try-on is a more elegant approach as it naturally facilitates the articulation of view, appearance and lighting etc. %
    % The research problem of
    
   3D garment retargeting aims at realistic draping of a 3D garment over 3D digital avatars of humans in varying shapes \& poses by inducing geometrical deformations (both rigid and non-rigid) over the garment surface, arising due to such variations. The problem of 3D garment retargeting is challenging because of several factors: arbitrary body shapes and poses, topological differences among various categories of garments, deformations arising out of the physical interaction with the underlying body, and resolving the penetration/intersection of the garment with the underlying body.
      
   % However, a fashion consumer on an AR/VR application would desire to see how a 3D garm would look when draped onto their 3D digital avatar. Thus,  asking the consumer to provide an animation trajectory for accurate retargeting is not feasible.
   
   % However, this has multiple challenges, such as handling 
    % \textbf{\textit{retargeting is a harder problem than simulation}}
    %Though the 3D space for tackling this problem is advantageous, the problem is still very \emph{ill-posed and requires addressing several challenges}, such as handling arbitrary body shapes and poses, modelling topological differences among various categories of garment
    % s along with realistic deformations arising out of the physical interaction with the underlying body, and resolving the penetration/intersection of the garment with the underlying body.
    % Virtual try-on technology has become increasingly popular in recent years due to its potential to improve the online shopping experience, reduce returns, and increase sales for retailers. Researchers have been exploring 2D virtual try-on methods; however, dealing in 2D image space limits the ability to model geometrical deformations correctly and provides a flat texture-based representation. On the other hand, 3D garment retargeting promises a more accurate representation as the draped garment can be viewed from different angles and perspectives, while catering to precise size-fitting, and can be easily extended to AR/VR environments for a more immersive experience and enhanced interactivity.  
    % large scale shape, pose and size variations,
    %  Recently,

        \begin{figure}[!h]
    \centering
        \includegraphics[trim={8cm 0 0 8cm},clip, width=\linewidth, trim]{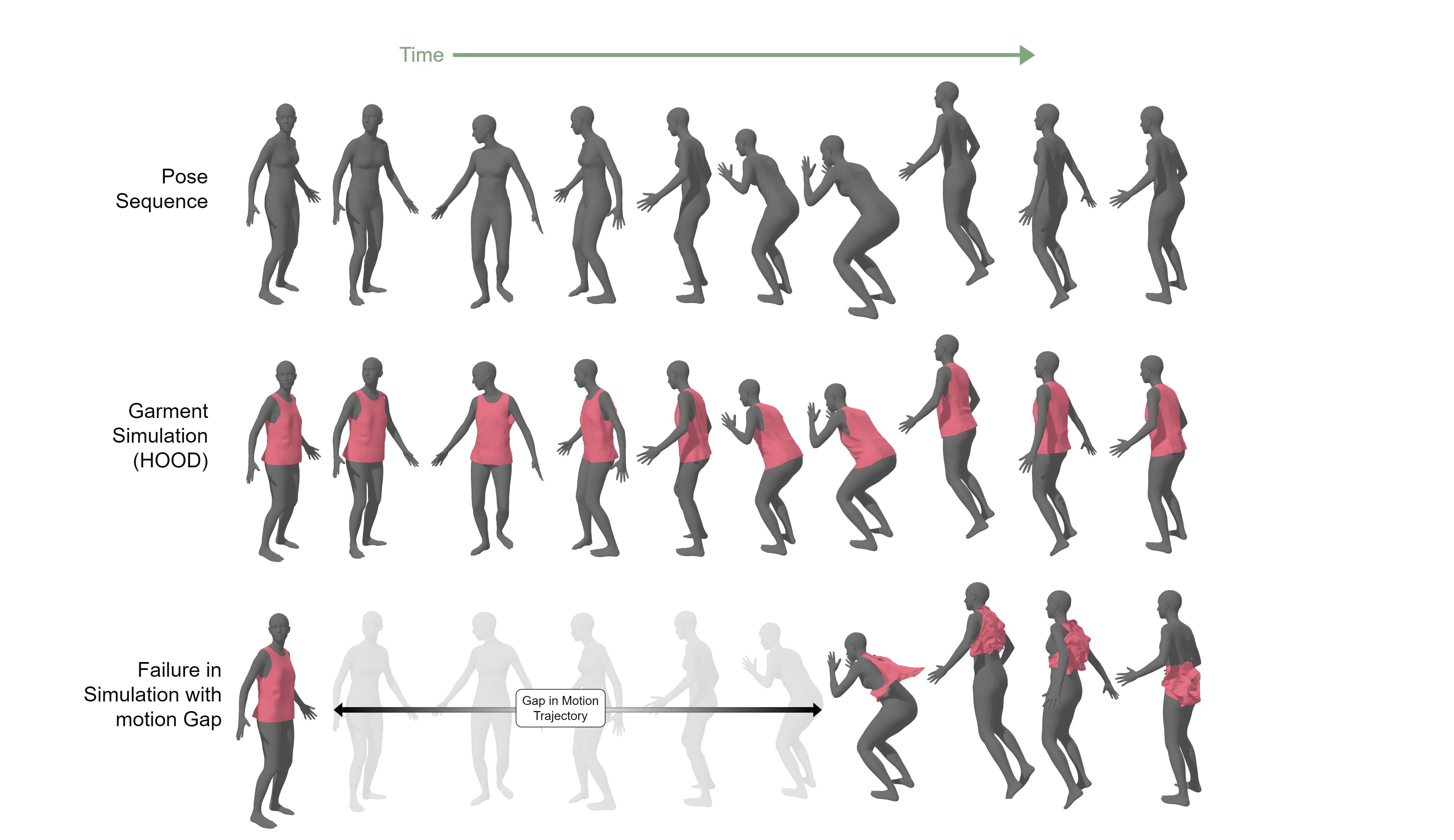}
        \caption{Failure of SOTA neural garment simulation-based methods to perform retargeting of the 3D garment from one arbitrary pose to the other when intermediate poses are unavailable.}
        \label{fig:hood_gap}
    \end{figure}
    
    Due to advancements in the field of deep learning and improvements in compute hardware over the past few years, researchers have been proposing various learning-based solutions to handle the problem of 3D virtual try-on. Parametric body models, such as SMPL\cite{SMPL:2015}, have made it easier to deal with the articulation of the human body and garments up to an extent. Various researchers have proposed \cite{patel20tailornet, scanimate, scale, POP:ICCV:2021} etc. which aim to model the dynamics of the garment draped on a parametric body as it changes. Recent developments in this direction have led to a plethora of self-supervised neural garment simulation approaches \cite{SNUG, neural_cloth_sim, HOOD}. At first glance, it looks like such methods have the capability to perform 3D garment retargeting. However, garment simulation deals with a fundamentally different scope, where the goal is to realistically deform the garment gradually as the underlying body dynamically changes the pose over an animated sequence. It assumes a complete trajectory of the underlying body going from an initial pose to a final pose. While these methods provide an accurate detailing of deformation and wrinkles in time by imposing physics-based constraints, they often rely on the previous frames to obtain simulation-specific parameters, e.g. velocity and acceleration information. Furthermore, direct retargeting (or simulating) the garment from one pose to another arbitrary pose fails drastically due to lack of motion information between the garment's pose and the target body pose (see Fig.\ref{fig:hood_gap}). Additionally, these methods do not support changing the shape/subject in between the simulation. On the other hand, garment retargeting deals with the transfer of a garment from one arbitrary pose to another, even on a different subject altogether. Methods, such as DIG\cite{dig} and DrapeNet\cite{drapenet} address this limitation by learning skinning weights to deform the garment from a canonical pose to any arbitrary pose in a self-supervised manner. However, to perform retargeting, the garment should be given either as a latent code of a large garment embedding space (learned using supervision\cite{drapenet}), or by fitting observations on a given image or 3D scan of the garment a latent template/code is retrieved which might not be a true representation of the garment mesh. Also, they cannot support draping the garment onto non-parametric human body.
    
    Recently introduced state-of-the-art work for 3D virtual-tryon, \cite{M3DVTON}, claims to propose the first 3D VTON solution by extending the 2D TPS-driven generative pipeline to reconstruct the 3D geometry, finally blending on a try-on image, with a representation similar to that of Moulding-Human \cite{mouldinghumans}. Though this approach allows viewing the draped garment on the target body from arbitrary viewpoints, the retargeting is still performed in 2D image space using a generative architecture and hence suffers from inherent limitations, e.g. blurry artifacts and false geometrical deformations. Additionally, since the method starts from the image of a garment, extending it to a real-world scan of a 3D garment is not possible.

    \begin{figure}[h!]
        \centering
        \includegraphics[width=0.48\textwidth]{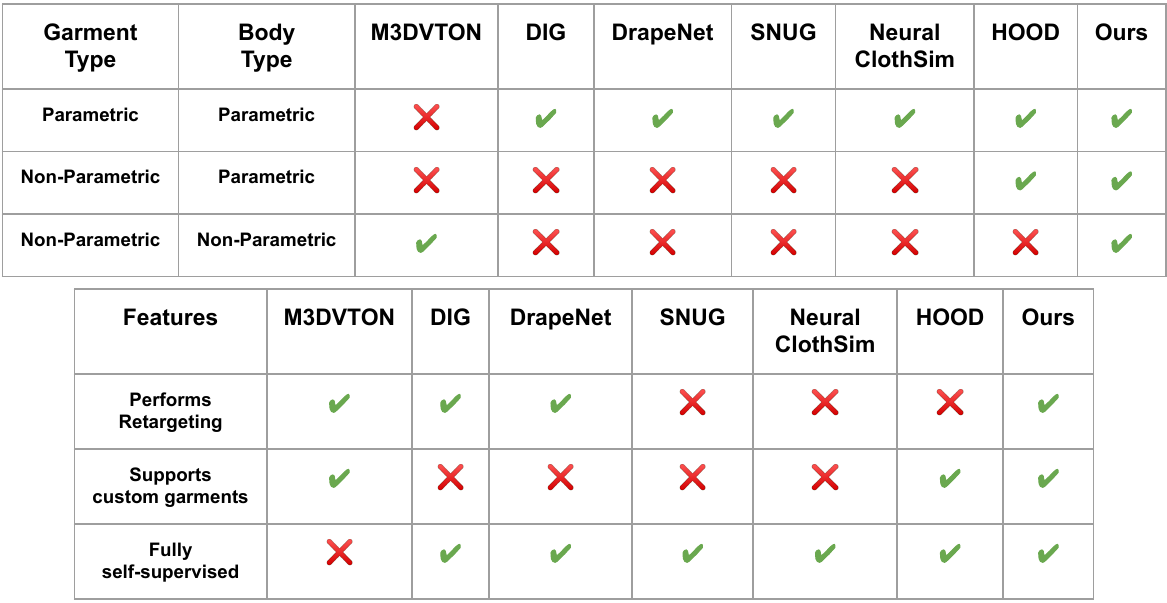}
        \caption{Compared to existing approaches, our proposed self-supervised garment retargeting method works for both parametric and non-parametric garments/bodies. }
        \label{fig:checklist}
    \end{figure}
    
In this work, we propose a robust, self-supervised method that can retarget real, parametric/non-parametric garment meshes over a target parametric/non-parametric human body, as shown in Fig.\ref{fig:teaser}. Given a 3D garment mesh and a target 3D human mesh, we first estimate correspondences between the two meshes using a novel representation, which provides an initial placement of the garment around the target body as a coarse retargeting initialization. We then employ a self-supervised training strategy, where we refine the coarse initialization and model shape and pose-specific deformations by minimizing the standard physics-based losses. Unlike existing methods \cite{dig, drapenet}, our framework doesn't learn skinning weights, therefore, can repose any arbitrary non-parametric garment on any parametric or non-parametric target body. 
Finally, as a post-processing step, we preserve the high-fidelity geometric details of the input garment and integrate it with the refined retargeted garment using \cite{sorkine2004laplacian}. The advantages of our proposed approach over limitations of existing approaches are shown in Fig.\ref{fig:checklist}.   Additionally, due to the lack of any real-world datasets for 3D garment retargeting, we curate our own dataset captured using a multiview Azure Kinect RGBD setup, containing different garments worn by multiple subjects in arbitrary poses. Our dataset serves as the ground truth for evaluating the proposed method for 3D garment retargeting. In summary, our main contributions are:
\begin{itemize}
    \item We develop a novel framework for retargeting arbitrary 3D garments on a given arbitrary target human body. Our method is the first one to enable retargeting of real, non-parametric garments over any arbitrary target body.
    \item We propose a novel representation for estimating correspondences between 3D garments and the target human bodies based on \textit{isomap embeddings} robust enough for arbitrary non-parametric garments.
    \item We propose a first-of-its-kind real-world 3D VTON dataset for evaluating our approach.
\end{itemize}
\noindent
We plan to release both the dataset and the code to further accelerate the research progress in this domain.\\
\\
\noindent
\textbf{\textit{Please refer to the supplementary draft for a comprehensive background \& literature survey, as well as supplementary video for a better visual understanding.}}

}
% % \input{sec/2_relatedWork}
\section{Method}
{
    \label{sec:method}
    \noindent
    % start with citing the pipeline figure
    Our proposed framework, outlined in Fig.\ref{fig:sysDiag}, has three key modules, namely, \textit{Correspondence-guided Coarse Retargeting}, \textit{Self-supervised Refined Retargeting, and Detail Preservation Module}. The input garment and the target body are fed to the first module to estimate dense correspondences between them, providing an initial coarse retargeting. Subsequently, our self-supervised refinement network refines the garment mesh geometry and introduces target body-specific surface deformations. Finally, geometrical details from the input garment are retained using Laplacian detail integration. 
    % In this section we will discuss and explain the methodology of our framework in detail. As discussed earlier, our framework does not require skinning weights and hence can retarget in-the-wild non-parametric 3D garment meshes with arbitrary topologies to any target body (SMPL mesh, real 3D human scan, digital mannequin etc.). One prominent feature of the proposed framework is that it can handle non-canonicalized garments (unlike \cite{}, which require garments generated via latent code to be in a canonical T-pose), i.e. a garment which is already draped on a 3D subject in a specific pose/shape/appearance can be easily retargeted to an entirely new subject with different pose/shape/appearance. To achieve this goal, we take a series of informed decisions, starting with correspondence-guided initialization (Sec.\ref{sub_sec:corres}); followed by modelling physics-based deformations for pose \& shape refinement (Sec.\ref{sub_sec:learning}); and finally, employing a wrinkle generation network as a post-processing module (Sec.\ref{sub_sec:wrinkle}).
    
    \subsection{Correspondence-Guided Coarse Retargeting}
    {
        \label{sub_sec:corres}
        \noindent
        % For draping 
        The aim of this module is to perform a coarse retargeting of the garment mesh over the target body mesh by first establishing dense surface-level correspondences between the two. Utilizing these correspondences, we transform the garment mesh vertices to align with the target body mesh vertices. The key idea is to establish dense correspondences which can provide a \emph{coarse} understanding of how the garment should be draped on the target body; e.g., sleeves going around the arms, the collar going around the neck etc. 
        %There is plenty of literature on correspondence-based 3D shape matching \cite{}, however, our aim is to use correspondences that capture a semantic understanding of the garment, while encapsulating the generic features of human bodies of all shapes and sizes. 
        SMPL\cite{SMPL:2015}, being a parametric body model, is a natural choice for acting as a medium for establishing dense surface correspondences, as it can easily model variations in human shapes \& poses. Therefore, we first perform dense non-rigid registration of both garment and target body mesh with the SMPL mesh, as shown in Fig.\ref{fig:sysDiag}. It is important to note that, unlike other methods \cite{dig, drapenet} which require perfectly registered SMPL mesh with the garment mesh, our approach can deal with noise in SMPL registration as we use it only to achieve initial coarse retargeting of garments.
        % (please refer to the supplementary material for more detail on SMPL registration). 

        % natural choice to expliot for the establishment of correspondences between the two meshes.?
        % For the task of draping a 3D garment onto a target 3D body, a natural thought that comes to mind is to establish some sense of correspondence between the two. The correspondences can provide a coarse understanding of how the garment should be draped on the target body; for e.g., sleeves should go around the hands, the collar should go around the neck etc. There is plenty of literature on correspondence-based 3D shape matching \cite{}, however, our aim is to use correspondences which capture a semantic understanding of the garment and the underlying human body. Moreover, these correspondences should encapsulate the generic features of garments and human bodies of all sorts. Therefore, with the objective of establishing generic human-centric correspondences, we choose to leverage SMPL\cite{SMPL:2015}, as it can easily model arbitrary human shapes \& poses.\\
                
        %\noindent
        %\textbf{Correspondence Estimation:} 
        % Subsequently, we extend dense surface correspondences to the garment mesh and target mesh by extrapolating from the registered SMPL mesh. 
        Let the garment mesh be $\mathcal{G}$, target body mesh be $\mathcal{T}$ and their corresponding  SMPL meshes be $\mathcal{M_G}$ \& $\mathcal{M_T}$, respectively. Establishing correspondences between $\mathcal{G}$ and $\mathcal{T}$ simply means for each vertex $v_i\in$ $\mathbb{R}^3$ of $\mathcal{G}$, locating a 3D point $x_i\in$ $\mathbb{R}^3$ on the surface of $\mathcal{T}$, where $v_i$ should be coarsely placed.
        % A naive approach to establish correspondence pairs $(v_i, x_i)$ from $\mathcal{G}$ to $\mathcal{T}$ is to choose the nearest vertex $q_i$ of $\mathcal{M_G}$ for each $v_i$ and the surface point $x_i$ of $\mathcal{T}$ located nearest to $q_i$ of $\mathcal{M_T}$ (for both $\mathcal{M_G}$ \& $\mathcal{M_T}$, index for $q_i$ is same, as the triangulation/connectivity of SMPL mesh is fixed.). However, the nearest-neighbor-based approach doesn't guarantee to model deformations far from SMPL mesh surface and is also highly susceptible to noise in the SMPL registration.
        One can perform simple skinning of the garment via the underlying SMPL mesh, but that only allows re-posing the garment into various poses and doesn't help in retargeting to different subjects. Alternatively, a naive way would be to find out the nearest SMPL vertex for the point on the garment and associate it with the corresponding nearest SMPL vertex to the human scan, but this approach produces a lot of local noise as an SMPL vertex can be associated to multiple garment/scan vertices (see Fig.\ref{fig:smpld}).
        \begin{figure}
            \centering
            \includegraphics[width=0.5\textwidth]{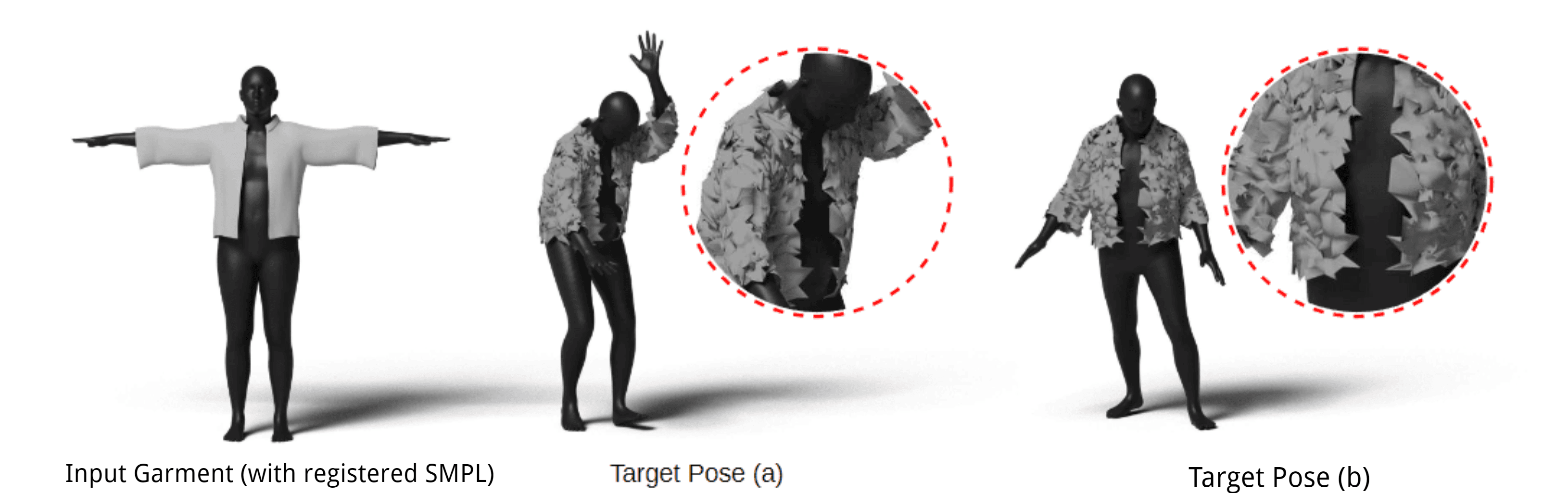}
            \caption{\textbf{SMPLD-based approach:} Naively using nearest neighbor among SMPL vertices results high-frequency local noise.}
            \label{fig:smpld}
        \end{figure}
        
        To mitigate the aforementioned issues and produce a locally smooth retargeting, we first define global features $\phi_i$ for each vertex $q_i$ of the SMPL meshes $\mathcal{M_G}$ \& $\mathcal{M_T}$. We later describe what feature space to use, but for now assume that we have predefined features for SMPL mesh vertices. we extrapolate these features to the vertices of $\mathcal{G}$ \& $\mathcal{T}$, and then perform correspondence matching based on these features. More specifically, the task is to estimate a feature vector $\phi_{smpl} = [\phi_{1}, \phi_{2},..., \phi_{6890}] \in \mathbb{R}^{6890 \times d}$ for each vertex $q_i$ of SMPL mesh, where $\phi_i \in \mathbb{R}^d$. $\phi_{smpl}$ is same for any SMPL mesh registered with any garment or body, i.e. $\phi_{smpl}=\phi_{\mathcal{M_G}}=\phi_{\mathcal{M_T}}$. Then, feature vector for each vertex $v_i$ of $\mathcal{G}$ is computed as follows:
        
        \begin{equation}
        \label{eq:phi_ext}
            \phi_{\mathcal{G}}^i = \frac{\sum_{j=1}^{k} [\phi_{\mathcal{M_G}}^j/dist(v_i, q_j)]}{\sum_{j=1}^{k} [1 / dist(v_i, q_j)]}; q_j \in \mathcal{N}^i
        \end{equation}
        \begin{equation}
            \mathcal{N}^i = [ q_1, q_2,..., q_k ]
        \end{equation}
        where, $dist()$ is the $\mathbb{L}_2$ distance, $q_j$ is a vertex of $\phi_{\mathcal{M_G}}$ \& $j^{th}$ nearest neighbor of $v_i$ in Euclidean space; and $|\mathcal{N}^i|=k=32$ (set empirically). Similarly, we compute $\phi_{\mathcal{T}}$ by extrapolating  $\phi_{\mathcal{M_T}}$ based on $k$-nearest neighbor distance.
        
        %Before making a suitable choice for $\phi_{smpl}$ to start with, few points need to be considered.
        Now, we describe what features to use for SMPL vertices and how to estimate them. Few essential aspects to be taken into consideration for choosing appropriate $\phi_{smpl}$.
        % Few important points needs to be considered before making a suitable choice for $\phi_{smpl}$.
        First, the feature embedding $\phi_{smpl}$ should incorporate both the local neighborhood information, while maintaining global structural context. Moreover, it should be concise yet representation-rich to uniquely characterize the associated surface, especially when extrapolating to the registered garment mesh or target body mesh. 
        %A poor choice would be to assign unique RGB colors to each vertex of the SMPL mesh and use them as features. This would result in a lot of repeated RGB values during extrapolation to the registered mesh, resulting in false correspondence matching (for e.g., hand and neck might get assigned very close RGB values, which might result in the sleeve of the garment mapping to the neck region). 
        Additionally, $\phi_{smpl}$ should be continuous over the surface of SMPL mesh to ensure locally smooth encoding of neighborhood information. We experimented with existing representations such as CSE\cite{ContinuousSE} and BodyMap\cite{bodymap} to serve the need for $\phi_{smpl}$, as they promise to encode global structural information. However, we empirically found them to produce false matching due to the repetition of extrapolated features due to very low dimensionality (we provide a detailed study regarding this in the supplementary).
        
        Thus, we develop a new strategy to establish correspondence across different garments and human body via SMPL, leveraging the intrinsic geometry-based Isomap Embeddings\cite{Jampani_2016_CVPR}. In order to encode local neighborhood information, we first compute the pairwise geodesic distance matrix, $|\mathbb{D}_{geo}| = 6890 \times 6890$, for all pairs of vertices $(q_i, q_j)$ of the SMPL mesh; i.e.
        \begin{equation}
        \label{eq:D_geo}
            {\mathbb{D}_{geo}}^{ij} = geodist(q_i, q_j)
        \end{equation}
        To incorporate global information, we use isometric mapping to fit the vertices of SMPL mesh onto a $d$ dimensional manifold by extending metric multi-dimensional scaling (MDS) based on $\mathbb{D}_{geo}$. This gives us a $d$-dimensional representation of each SMPL vertex $q_i$, i.e. $\phi_{smpl}$. We empirically found that setting $d$=128 ensures sufficient dimensionality to avoid repetitions {\color{black}while extrapolating} on the target or registered mesh. Finally, we estimate $\phi_{\mathcal{G}}$ \& $\phi_{\mathcal{T}}$ using Eq.\ref{eq:phi_ext}. These extrapolated features are termed as \textit{\textbf{Isomap Embeddings}}.
        
        %\noindent
        %\textbf{Initial Coarse Retargeting:} 
        Based on the estimated \emph{ Isomap embeddings}, we first perform an initial retargeting to \emph{coarsely} position the garment around the target body. In particular, for each vertex $v_i$ of $\mathcal{G}$, the corresponding 3D target location $x_i$ in the vicinity of $\mathcal{T}$ is estimated as follows:
        \begin{equation}
        \label{eq:knn_corres}
            x_i = \frac{\sum_{j=1}^{k} [u_j/dist(\phi_{\mathcal{G}}^i, \phi_{\mathcal{T}}^j)]}{\sum_{j=1}^{k} [1 / dist(\phi_{\mathcal{G}}^i, \phi_{\mathcal{T}}^j)]}; \phi_{\mathcal{T}}^j \in \mathcal{N}^i
        \end{equation}
        \begin{equation}
            \mathcal{N}^i = [ \phi_{\mathcal{T}}^1, \phi_{\mathcal{T}}^2,..., \phi_{\mathcal{T}}^k ]; \phi_{\mathcal{T}}^j \in \phi_{\mathcal{T}}
        \end{equation}

        where, $dist()$ is the $\mathbb{L}2$ distance,$u_j$ is the vertex of target mesh $\mathcal{T}$ corresponding to $\phi_{\mathcal{T}}^j)$ , $\mathcal{N}^i$ the set of $k$-nearest neighbors of  $\phi_{\mathcal{G}}^i$ in $\phi_{\mathcal{T}}$, and $|\mathcal{N}^i|=k=32$. We replace the vertices $v_i$ of $\mathcal{G}$ with corresponding $x_i$, coarsely retargeting the garment mesh around the target mesh $\mathcal{T}$. Fig.\ref{fig:sysDiag}(e) \& (f) gives a visual overview of this process. For an arbitrary point on the garment, an initial target 3D point on the target is located via \emph{Isomap Embedding vectors}.
        This coarse initialization is then refined using a self-supervised strategy explained in the next section.

    % \subsection{Self-Supervised Refinement}
    % {
    %     \label{sub_sec:physics_based}
    % }

    % \begin{figure*}
%     \centering
%     \begin{minipage}{\linewidth}
%         \centering
%          % \includesvg[width=\textwidth]{assets/images/smpl.png}
%          \includegraphics[width=\textwidth]{assets/images/smpl.png}
%     \end{minipage}
%         \caption{\emph{System overview}: We capture a 3D scene of voxelized radiance field and distill the semantic feature into it. Once captured, the user can easily mark a regions using a brush tool on a reference view (green stroke). The features are collected corresponding to the marked pixels and clustered using K-Means. The voxel-grid is then matched using NNFM (nearest neighbor feature matching) to obtain a high confidence seed using a common tight threshold. The seed is then grown using bilateral search to smoothly cover the boundaries of the object, conditioning the growth in the spatio-semantic domain.}
%     \label{fig:sysDiag}
% \end{figure*}
 \begin{figure*}
        \centering
        %\begin{minipage}{\linewidth}
        \includegraphics[width=0.66\linewidth, trim={1cm 0 0 0},clip]{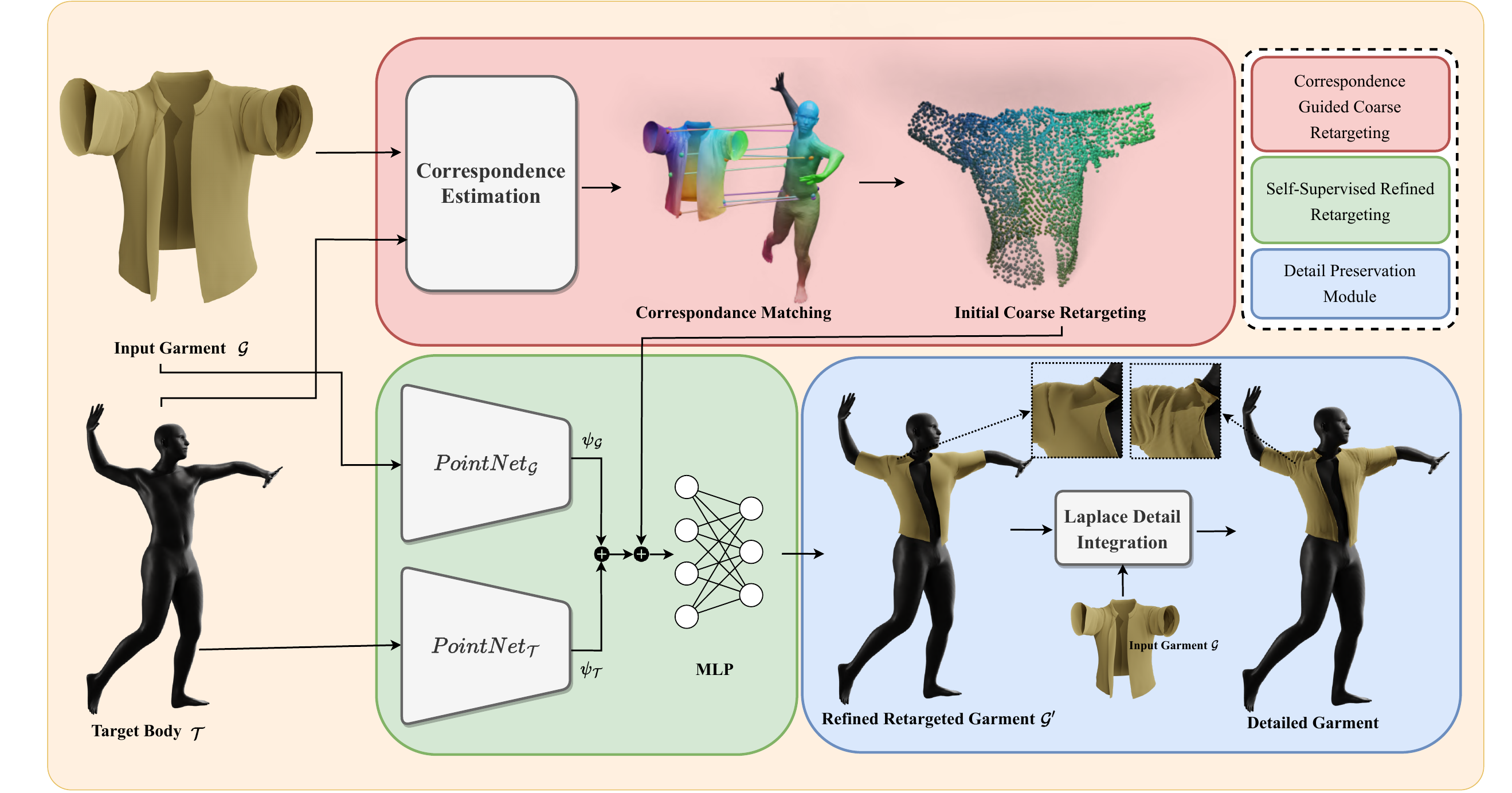}\includegraphics[width=0.36\linewidth]{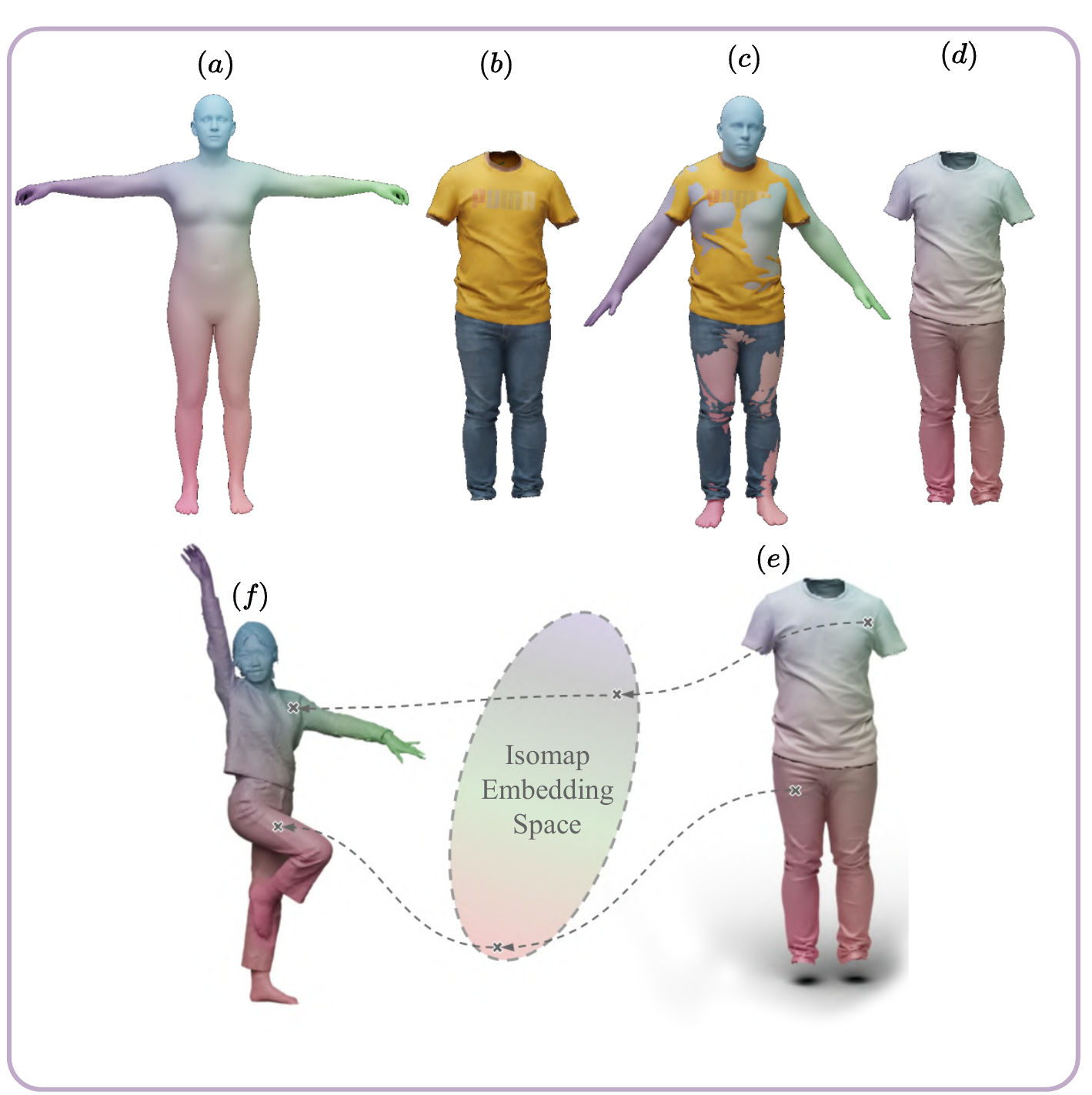}        %\end{minipage}
        \caption{\emph{Outline of the proposed self-supervised garment retargeting framework \textbf{(left)}; and visualization of Isomap embedding estimation for arbitrary 3D scans.\textbf{(right)}: (a) SMPL mesh with per-vertex Isomap embeddings; (b) Input 3D garment(s); (c) SMPL registered with the input garment(s); (d) Isomap embeddings transfered to the input garment.}.}
        \label{fig:sysDiag}
\end{figure*}

    }
    
    % \subsection{Global Context}
    % {
    %     \label{sub_sec:global}
    %     The task of draping garment onto a novel pose and shape is ill-posed problem as discussed in Sec.\ref{sec:intro}. Learning a distribution over varied body poses and profiles is desirable solution. In recent times the neural fields\cite{neural_fields} have shown promising results in the usage of simple-MLP architectures to produce succinct results for learning such distributions. Inspired by these works we adopted MLPs to produce desired output on varied poses and sizes of the body. When the MLP is only fed with positional information the network not able to generalize across pose and shape variations of human body. To address this we condition the network contingent to the SMPL shape changes. We do a layerwise conditioning of MLP with the help of PointNet features, which provide global context to our network. We observed a generalization in the network performance in predicting the residual and refinement for pose and shape variations. (Refer Fig.\ref{fig:sysDiag})
            
    %     For shape understanding we also feed IsoMap embeddings of point of interest to the network which share a common space across the human body. Since isomap embeddings are independent of the pose, body profile they contribute in providing global context to the network. For calculation of these embeddings we rely on SMPL. Do note that, while it is true that we leverage SMPL for these embeddings, we do not base our method on blend-weights which restricts the modeling. 
    % }

    \subsection{Self-Supervised Refined Retargeting}
    
    {
        \label{sub_sec:learning}
        Given a coarsely retargeted garment mesh, where the garment vertex mesh coordinates $v_i$ are replaced by their respective correspondence surface points $x_i$ on target body mesh, we propose to refine these vertex positions further to incorporate accurate pose \& shape-specific deformations. However, supervised learning is not suitable for this refinement task due to the lack of ground truth pairs on real data. Thus, we resort to a self-supervised setup where we minimize losses that try to maintain the original topology of the garment mesh (namely, retaining edge lengths and relative face orientation) while preserving the coarse retargeting. 
        
        %\noindent
        Let the refined vertex positions of the garment mesh $\mathcal{G'}$ be $v'_i = x_i + \Delta x_i$. We employ a Multi-Layer Perceptron (MLP)  network to predict per-vertex $\Delta x_i \in \mathbb{R}^3$. The per-vertex input to the MLP is $\mathcal{I} = \{ x_i\ ,\ \phi_{\mathcal{G}}^i\ ,\ \chi^{k,i}_{\mathcal{M_T}}\ ,\ \psi_{\mathcal{G}}\ ,\ \psi_{\mathcal{T}}\ \}$. Here, $x_i \in \mathbb{R}^3$ is $i^{th}$ vertex-position of the coarsely retargeted mesh and $\phi_{\mathcal{G}}^i \in \mathbb{R}^{128}$ is the corresponding isomap embedding. Additionally, the MLP also takes $k$-nearest neighbours of $x_i$ belonging to the vertex set of target body mesh $\mathcal{T}$, denoted as $\chi^{k,i}_{\mathcal{M_T}}$ $(k = 32)$. In order to encode a useful global context for both garment and target body, we use two separate PointNet\cite{PointNet} encoders, which provide $128$ dimensional global encoding of the vertices of the garment mesh and the body mesh, denoted as $\psi_{\mathcal{G}}=PointNet_{\mathcal{G}}(vertices(\mathcal{G}))$ \& $\psi_{\mathcal{T}}=PointNet_{\mathcal{T}}(vertices(\mathcal{T}))$, respectively. Both the encoders are trained jointly with the MLP decoder in a self-supervised fashion to minimize the following losses:\\

        \noindent\textbf{Edge-Length loss:} This loss is used to preserve the structural integrity of the garment by constraining the change in the length of the edges of the original garment mesh, calculated as follows:
            \begin{equation}
                \mathcal{L}_{length} = \frac{1}{m}\sum_{i=1}^{m} w_i \cdot \left\Vert e_i - e'_i \right\Vert
            \end{equation}
            \begin{equation}
              w_i =
              \begin{cases}
                0 & \text{if $e_i \in \mathbf{J}$} \\
                1 & \text{otherwise}
              \end{cases}
            \end{equation}
            where, $e_i \in edges(\mathcal{G})$, $e'_i \in edges(\mathcal{G'})$ and $m = |edges(\mathcal{G})|$. $\mathbf{J}$ is the set of edges of the garment mesh belonging to the special joint locations of the underlying human body, specifically, elbows, armpits, waist, and knees (refer supplementary for details). These are the prominent regions that undergo extreme deformation due to pose change. Hence, we chose not to preserve edge length around such regions to allow accurate reposing of the garment.\\

         \noindent \textbf{Correspondence Loss:} Edge-length loss has the effect of retaining the original pose \& shape of the garment in order to maintain its structure. We employ an additional loss to constrain this behavior by ensuring that the correspondences between the refined garment and the target body should be similar as for the original garment used for coarse retargeting. The predicted residual $\Delta x_i$ is used to get refined vertex positions $v'_i \in \mathcal{G}$. We then compute correspondences $x'_i$ for each $v'_i$ using Eq.\ref{eq:knn_corres} and minimize the  $\mathbb{L}2$ norm between $x_i$ \& $x'_i$, i.e.
            \begin{equation}
                \mathcal{L}_{corres} = \frac{1}{n} \sum_{i=1}^{n} \left\Vert x_i - x'_i \right\Vert;\ n = |vertices(\mathcal{G})|
            \end{equation}
            It ensures that the garment doesn't deviate too much away from the initial coarse retargeting and remains in the vicinity of the target body.\\
            
         \noindent \textbf{Bend Loss:} We impose bend loss, introduced in \cite{SNUG}, to ensure that the angle between two adjacent faces is as low as possible. This makes sure that the output is smooth and does not have any weird deformations or artifacts.

    }
    \subsection{Detail Preservation Module}
    {
    Our self-supervised networks accurately refine the initial coarse retargeting in-order to retarget the input garment onto the given body. However, it tends to produce a smooth surface lacking high-frequency details of the garment (collars, pockets, etc.). Inspired by \cite{sorkine2004laplacian},  we preserve the high-fidelity geometric details of the input garment and integrate it with the refined retargeted garment.
    % we aim to preserve the high-frequency details of the input garment and integrate it with the refined retargeted garment.
        \begin{figure}[!h]
        \centering
        \includegraphics[width=0.8\linewidth]{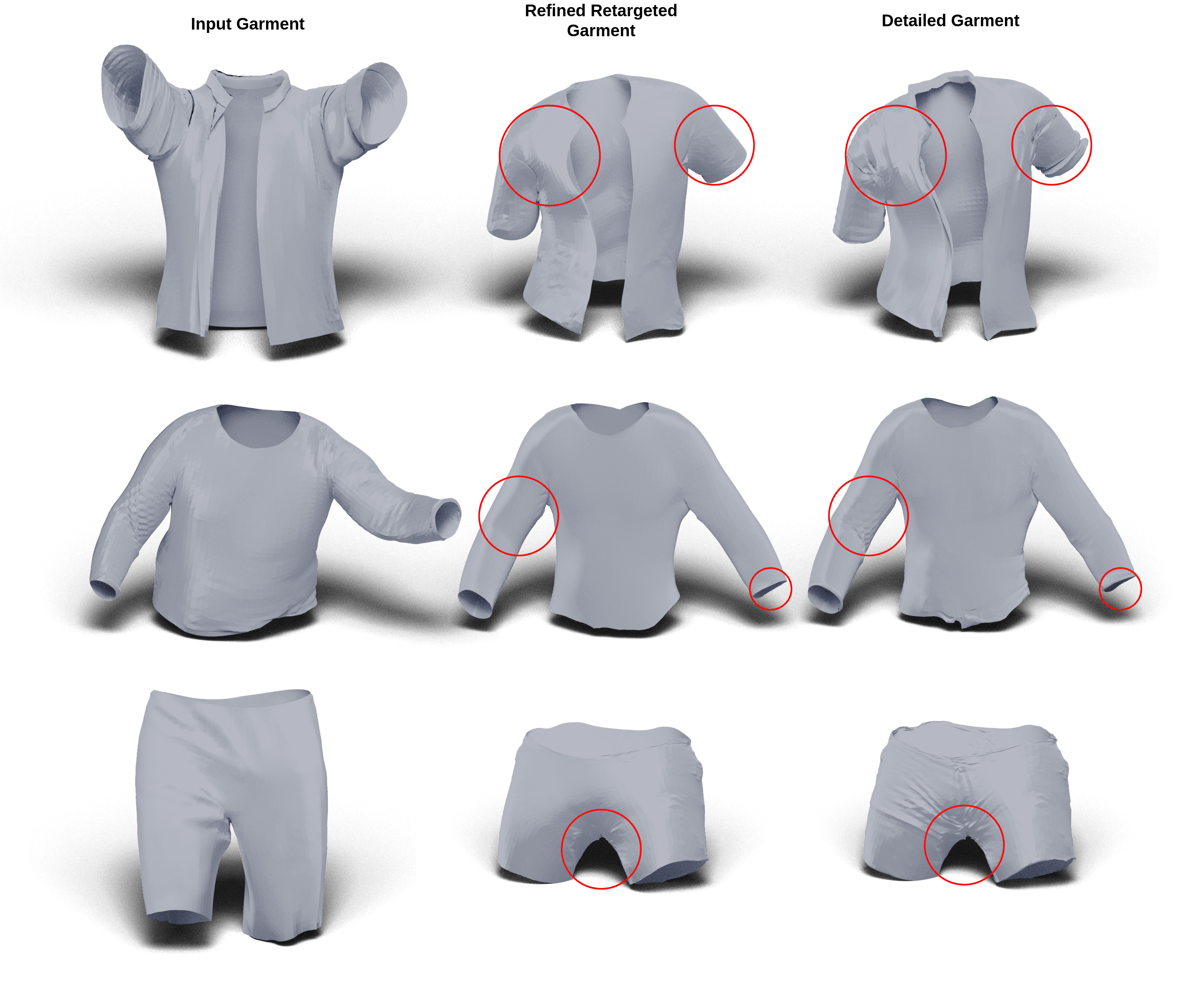}
        \caption{Results of Laplace detail integration. }
        \label{fig:detail_transfer2_comp}
    \end{figure}
    Given the input garment mesh $\mathcal{G}$ with $V_{\mathcal{G}}=\{v_1, v_2, .... v_N\}$ vertices in $\mathbb{R}^3$ where $\mathcal{N}$ is the total number of vertices
    the Laplacian Matrix can be used to retrieve the high fidelity details of the mesh. For each vertex $v_i$ let, $\mathcal{N}_i=\{j|(i,j) \in K\}$ be the neighborhood ring
    directly connected to $v_i$ and degree $d_i$ be the number of vertices in $\mathcal{N}_i$. The uniform Laplacian coordinate per vertex is given as:
    \begin{equation}
            \delta_i(v_i) = v_i - \frac{1}{d_i}  \sum_{j \in \mathcal{N}_k} v_j
    \end{equation}
    The above equation can be represented in matrix form: $L[v_1,v_2,...v_N]^T = [ \delta_1,\delta_2,....,\delta_N ]^T$ where $L$ is the uniform Laplacian Matrix given as
    $L=I-D^{-1}A$. Here $A$ is the mesh adjacency matrix and $D=diag(d_1,d_2...d_N)$ be the degree matrix. 
    
    In order to integrate the high-fidelity geometric details from input garment on to retargeted garment, we first calculate the uniform Laplacian Matrix $L_\mathcal{G}$ and
    Laplacian coordinates $\delta_\mathcal{G}$ of the input mesh $\mathcal{G}$. We fix anchor points on the retargeted mesh $\mathcal{G^{'}}$ and recompute the Laplacian 
    matrix as $\hat L = [L_\mathcal{G}^T,1_i]^T$ and Laplacian coordinates as $\hat \delta = [\delta_\mathcal{G}, v_i]^T$. $1_i$ is the one hot encoding where $i_{th}$ is 
    one. We finally obtain the retargeted mesh with high fidelity details $\mathcal{G^{''}}$ with $V_{\mathcal{G^{''}}}$ vertices by solving a linear system to obtain the 
    modified vertex positions as $V_{\mathcal{G^{''}}} = \hat L^{-1} \hat \delta$. We show the result of Detail Preservation module in Fig. \ref{fig:detail_transfer2_comp}
    }
}
\section{Experimentation \& Results}
{   

    %In this section, we discuss implementation details, datasets \& metrics used for evaluation, and report qualitative and quantitative results along with appropriate ablations.
    \subsection{Implementation Details}
    \label{sub_sec:impl_details}
    {
        For the establishment of correspondence-based retargeting, we utilize open frameworks like Trimesh and Open3D. For the self-supervised refinement of retargeting we utilize an MLP based model. The MLP consists of 512 neurons per layer and has 6 such layers with 6 layers. The MLP is fed with PointNet encodings of the SMPL and garment mesh\cite{PointNet} along with every point $x$ of coarse retargeted body. We implement this interface in PyTorch. Additional implementation \& training details are mentioned in the supplementary document.
    }
    
    \subsection{Datasets}
    {
        \label{sub_sec:impl_details}
        To evaluate our approach, we require ground truth 3D garments to be draped over the target body of poses and shape variations. %However, as mentioned earlier, there is a significant lack of such large-scale datasets. 
        CLOTH3D \cite{bertiche2020cloth3d} is the only dataset that offers data in the required setting. However, the garments are synthetic and parametric in nature, draped using a simulated engine. Hence the lack of real-world aesthetics and noise is prevalent. To address this gap, we capture our own dataset "DressMeUp" to validate our approach on a real-world data distribution. We briefly describe both datasets, and additional details are present in the supplemental document.
        
        %\noindent\textbf{AMASS:}
        %    
        \noindent\textbf{CLOTH3D:} Cloth3D provides a simulated collection of sequences containing clothed humans, modeled using SMPL meshes and their corresponding parametric garments. They model the animations in accordance to a large collection of MoCap data. The dataset offers a wide garment range(t-shirts, tank-tops, trousers etc.) which we broadly group into two categories $-$ TopWear \& BottomWear. 
    
        \noindent\textbf{DressMeUp (Our Dataset):} As stated earlier in Sec. \ref{sec:intro}, there is a need for real-world 3D garment datasets to validate the proposed methodologies, which contain realistic garments draped on real humans. To bridge this gap we captured around $\sim255$ meshes of real garments draped onto humans of varied poses and body profiles. We believe that this dataset provides a more rigorous evaluation, extending beyond the parametric modeling of clothing \& latent garments. 
        
        This data was captured using Azure Kinect-based multiview RGBD capture setup. We collected $\sim255$ garments scans, worn by 15 unique subjects, with 44 unique garments. {For every garment, a subject is scanned in 5 different poses.} Each pose is captured using a static multi-view(7) RGBD system. To obtain final mesh reconstructions we employ multiview Kinect Fusion\cite{kinectfusion} on the captured RGBD data. {\color{black}To further rectify the noise of the raw scan,  manual post-processing is performed utilizing the eclectic and elegant toolkit of Meshlab. While post-processing we also obtain a UV-mapped mesh of the garment to facilitate texture swapping.} Additionally, we perform SMPL registration for each mesh to approximate the pose \& shape. Our dataset captures realistic noise \& topological deformations of real-world garments draped over different subjects under different poses. We believe our dataset can prove to be extremely useful in the progress of the 3D-VTON domain.    
    }
    % \vspace 1mm

    \subsection{Evaluation Metrics}
    {  
        \label{sub_sec:eval_metrics}
        To quantitatively evaluate our proposed approach, we report widely used metrics like Euclidean Distance(ED), Normal Consistency(NC), Interpenetration Ratio(IR) and Point-to-Surface Distance(P2S). Please refer to the supplementary material for more details about these metrics.
    }

\begin{figure}
    \centering
    \includegraphics[width=\linewidth]{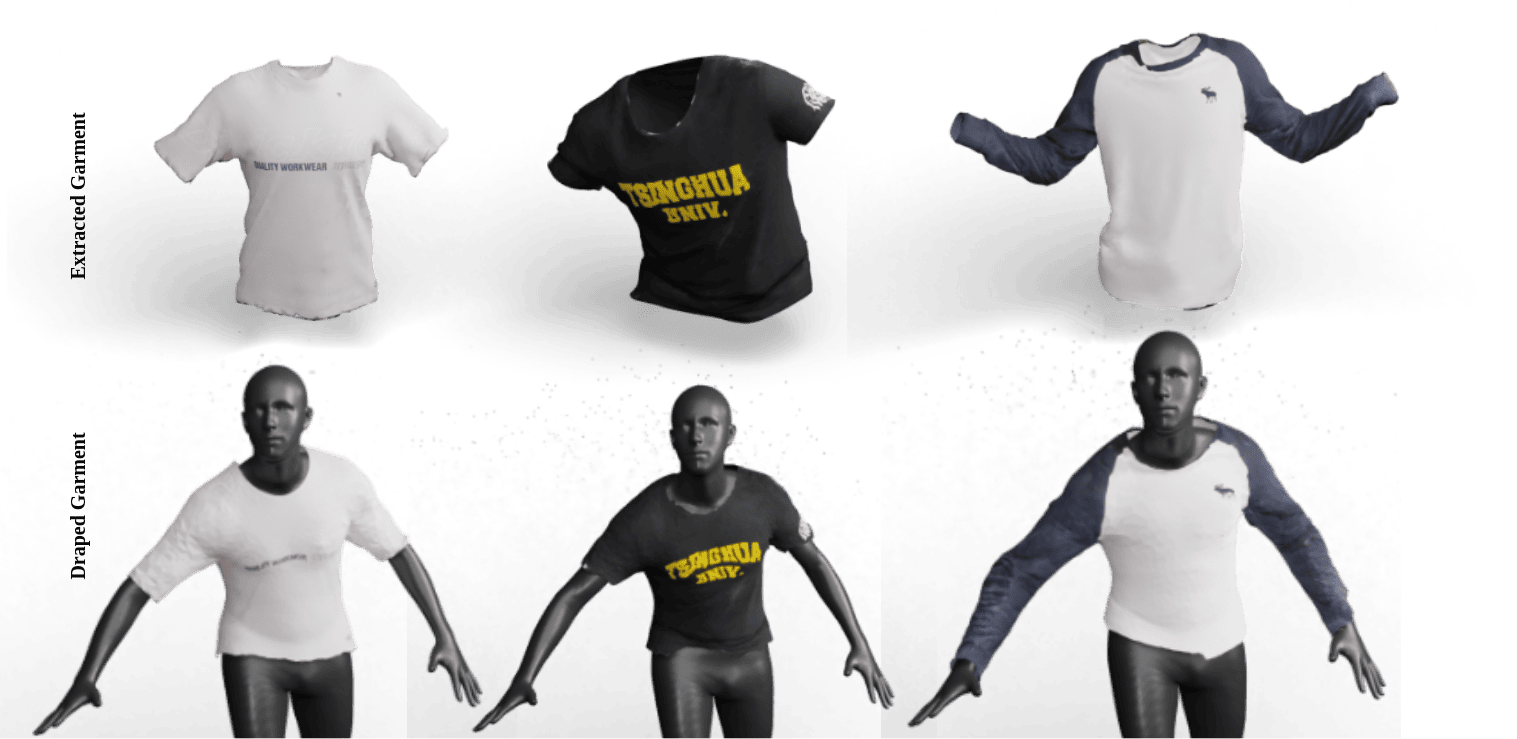}
    \caption{Results of real garments draped on unseen pose/shape.} 
    \label{fig:real_garments}
\end{figure}
    % \begin{figure*}[t]
%     \centering
%     \begin{minipage}{0.24\linewidth}
%         \centering
%         \subcaptionbox{Original Rendered Image}{\includegraphics[width=\textwidth, height=3cm]{assets/images/supp_editing/og.png}}
%     \end{minipage}
%     \begin{minipage}{0.24\linewidth}
%         \centering
%         \subcaptionbox{Removal of Pot}{\includegraphics[width=\textwidth, height=3cm]{assets/images/supp_editing/empty.png}}
%     \end{minipage}
%     \begin{minipage}{0.24\linewidth}
%         \centering
%         \subcaptionbox{Composition}{\includegraphics[width=\textwidth, height=3cm]{assets/images/supp_editing/lego.png}}
%     \end{minipage}
%     \begin{minipage}{0.24\linewidth}
%         \centering
%         \subcaptionbox{Style Transfer}{\includegraphics[width=\textwidth, height=3cm]{assets/images/supp_editing/style.png}}
%     \end{minipage}
%     \caption{\emph{Seamless Progressive Scene Editing}:
%     {Image (a) is the reference rendered viewpoint. In (b), the pot has been removed. Image (c) shows scene composition. The JCB from \kitchen scene has been placed on the top of the table in the \garden scene. Image (d) shows appearance editing of specific objects. We apply style transfer on just the JCB. For more details please refer to Sec. \ref{sec:sup_scene_editing}.}}
%     \label{fig:sup_editing}
% \end{figure*}

\begin{figure}
    \centering
    \includegraphics[width=\linewidth, trim={1.2cm 0 0 0},clip]{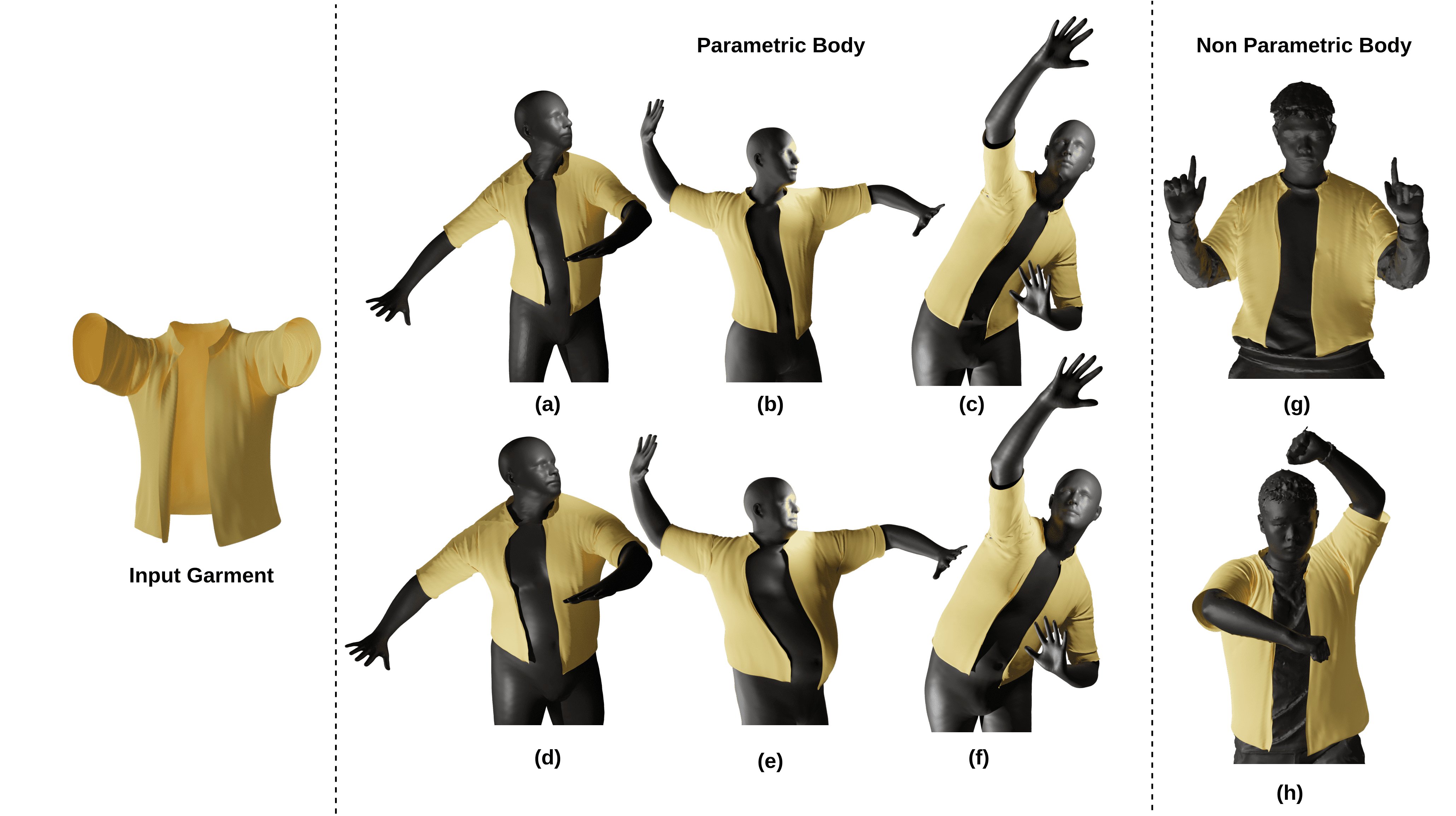}
    \caption{Results from our method for retargeting 3D garment onto SMPL body meshes of different poses and shapes (a) - (f); and on non-parametric 3D human scans (g) \& (h).}
    \label{fig:different_pose_shapes}
\end{figure}

\begin{figure} 
\centering
\includegraphics[width=1.1\linewidth]{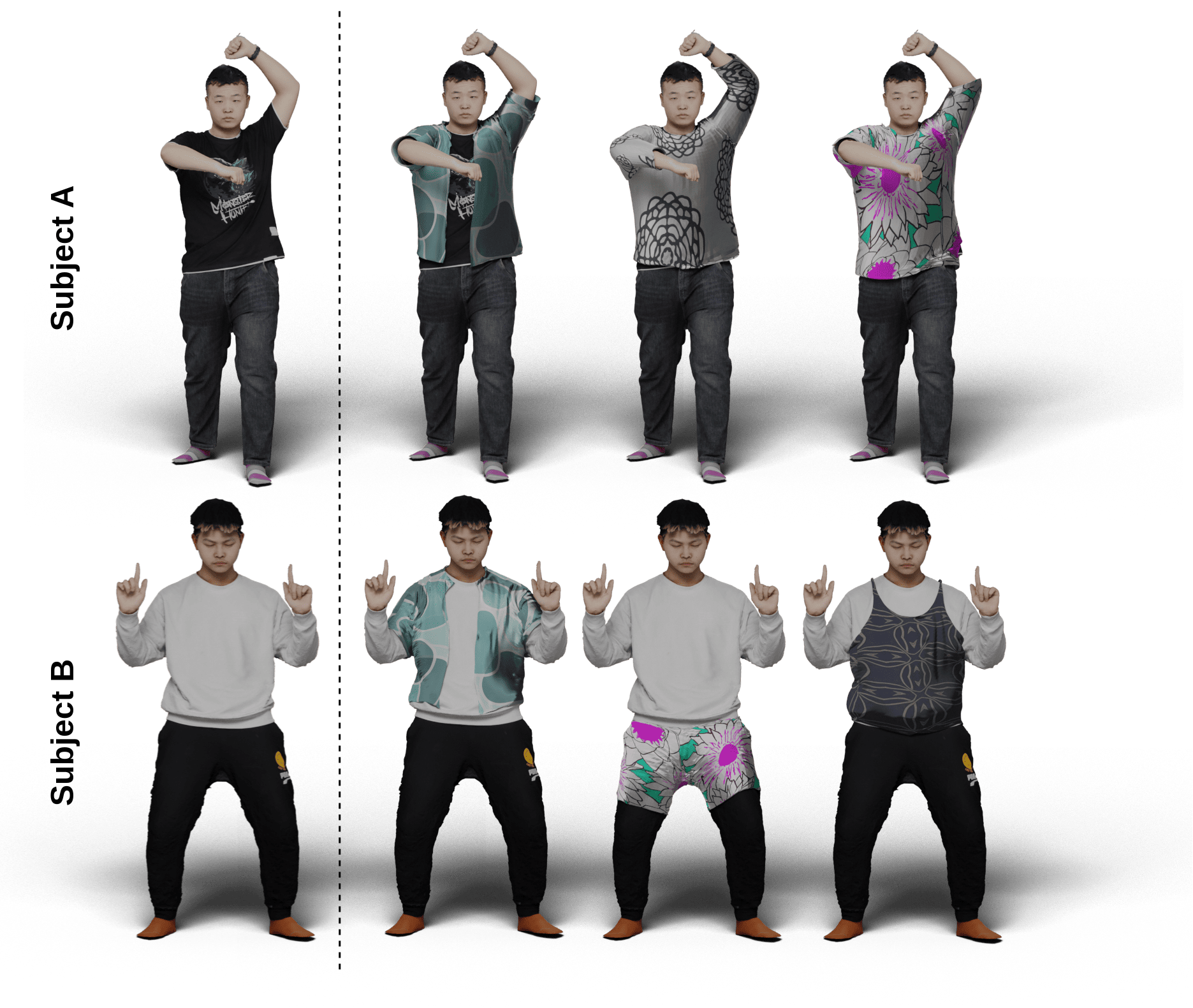}
\caption{Retargetting 3D garments from CLOTH3D dataset onto non-parametric human bodies from THumans2.0 \cite{thuman} dataset. Our approach can deal with layered clothing as well.} 
\label{fig:3dhuman_thuman_new}
\end{figure}
    \subsection{Results}
    \noindent\textbf{Qualitative \& Quantitative Results on CLOTH3D:} For evaluation purposes, we randomly select $\sim$273 random sequences from the CLOTH3D dataset. We uniformly sample 5 frames per sequence, ensuring that there is a significant \emph{pose} change among the sampled frames. Out of five sampled frames, we take SMPL bodies from the first three for self-supervised training and use the remaining for evaluation. Additionally, instead of taking garments from each sequence, we \emph{only} sample  10 garments out of the available corpus of garments for self-supervised training, to ensure evaluation is only done on unseen garments. Fig.~\ref{fig:different_pose_shapes} shows qualitative results of our framework on CLOTH3D dataset, where we report retargeting results on three different poses along with three different shapes. Our framework can retarget arbitrary unseen garments on the target bodies with varying poses and shapes, as evident in the figure. We also report quantitative metrics mentioned in Sec.\ref{sub_sec:eval_metrics} on the evaluation samples of CLOTH3D in Table.\ref{table:cloth3d_kinet_data}. We achieve sufficiently low ED, P2S, and IR metrics while maintaining high Normal Consistency.
        
     \noindent\textbf{Qualitative \& Quantitative Results on Our Dataset:}
        For evaluation of our dataset, we perform self-supervised training on $500$ target SMPL meshes from AMASS dataset to ensure enough pose variation, minimizing losses while learning to drape $10$ synthetic garments from CLOTH3D dataset. Even being trained on synthetic garments, our network is able to generalize on real garments from our dataset. Table.\ref{table:cloth3d_kinet_data} reports corresponding evaluation metrics where we achieve satisfactory performance. The values reported on CLOTH3D are slightly better because training and evaluation are both done on synthetic garments. However, in the case of our dataset, training is done on synthetic garments and evaluation on real garments, thereby leaving a window for an out-of-distribution scenario.
        
    \noindent\textbf{Qualitative Results on Real Scans:}
        Fig.~\ref{fig:real_garments} shows qualitative results of our framework on real garments retargeted to arbitrary SMPL meshes, and Fig.~\ref{fig:3dhuman_thuman_new} shows qualitative results on real target human scans. It is evident from both the figures that even being trained on synthetic garments and target SMPL meshes, our framework can retarget real garments on arbitrary real scans (not just SMPL meshes). This shows the generalization capabilities of our framework on real-world samples. We can also drape garments on top of other garments, making way for layered clothing as well.
        
    \noindent\textbf{Qualitative Results on Internet Images:}. Fig.~\ref{fig:internet} shows qualitative results of retargeting 3D garments onto 3D human meshes reconstructed from images (using \cite{ECON, reef}). This is yet another proof of good generalization of our method on in-the-wild OOD samples (e.g. yoga pose).
    \begin{table}
    \centering
    \begin{tabular}{|l |l l| l l|}
    \hline
    \multicolumn{5}{|c|}{\textsc{Cloth3D}} \\
    \hline
    {\textsc{Type}} & {P2S$\downarrow$} & \multicolumn{1}{c|}{ED$\downarrow$} & {NC$\uparrow$}  & {IR\%$\downarrow$}\\
    {}    & \multicolumn{2}{c|}{x $10^{-3}$} &  \multicolumn{1}{c}{} & \\
    \hline
    % \multicolumn{5}{|c|}{}\\
    {topwear}    & {6.901} & {9.353}  & {0.951} & {0.009} \\
    {bottomwear}   & {8.049} & {9.832}  & {0.943}  & {0.006}\\
    \hline
    % \multicolumn{5}{|c|}{}\\
    \multicolumn{5}{|c|}{\textsc{Our Captured Data}}\\
    \hline
    % \multicolumn{5}{|c|}{}\\
    {topwear} & {12.119} & {12.571}  & {0.854} & {0.037}\\
    {bottomwear} & {6.753} & {7.314}  & { 0.849} & {0.014}\\
    \hline
    \end{tabular}
    \caption{Quantitative evaluation/ benchmarking of our method on Cloth3D and  our Dress-Me-Up data.}
    \label{table:cloth3d_kinet_data}
\end{table}

% {\sleeveless} & { 6.3216} & {9.1641} & {0.2609}& {0.9665}  \\
% {\sleeved}    & {7.4813} & {9.5436} & {0.5925} & {0.9343}  \\
% \multicolumn{1}{|c}
% \multicolumn{1}{c|}
     \begin{table}
    \centering
    \begin{tabular}{|l |c c| c c|}
    \hline
    \multirow{2}{*}{Noise} & {P2S$\downarrow$} & \multicolumn{1}{c|}{ED$\downarrow$} & {NC$\uparrow$} & {IR\%$\downarrow$}\\
                           & \multicolumn{2}{c|}{x $10^{-3}$} & &\\
    \hline
    {$10^{-4}$} & {7.481} & {9.544}  & {0.934} & {0.009}\\
    
    {$10^{-3}$} & {7.521} & { 9.581}  & {0.927} & {0.009}\\
    
    {$10^{-2}$} & { 10.247} & {11.97}  & {0.761} & {0.014}\\
    \hline
    \end{tabular}
    \caption{Ablation regarding noise in correspondence estimation.}
    \label{table:noise_ablation}
\end{table}
    \begin{figure}
    \centering
    \includegraphics[width=0.6\linewidth]{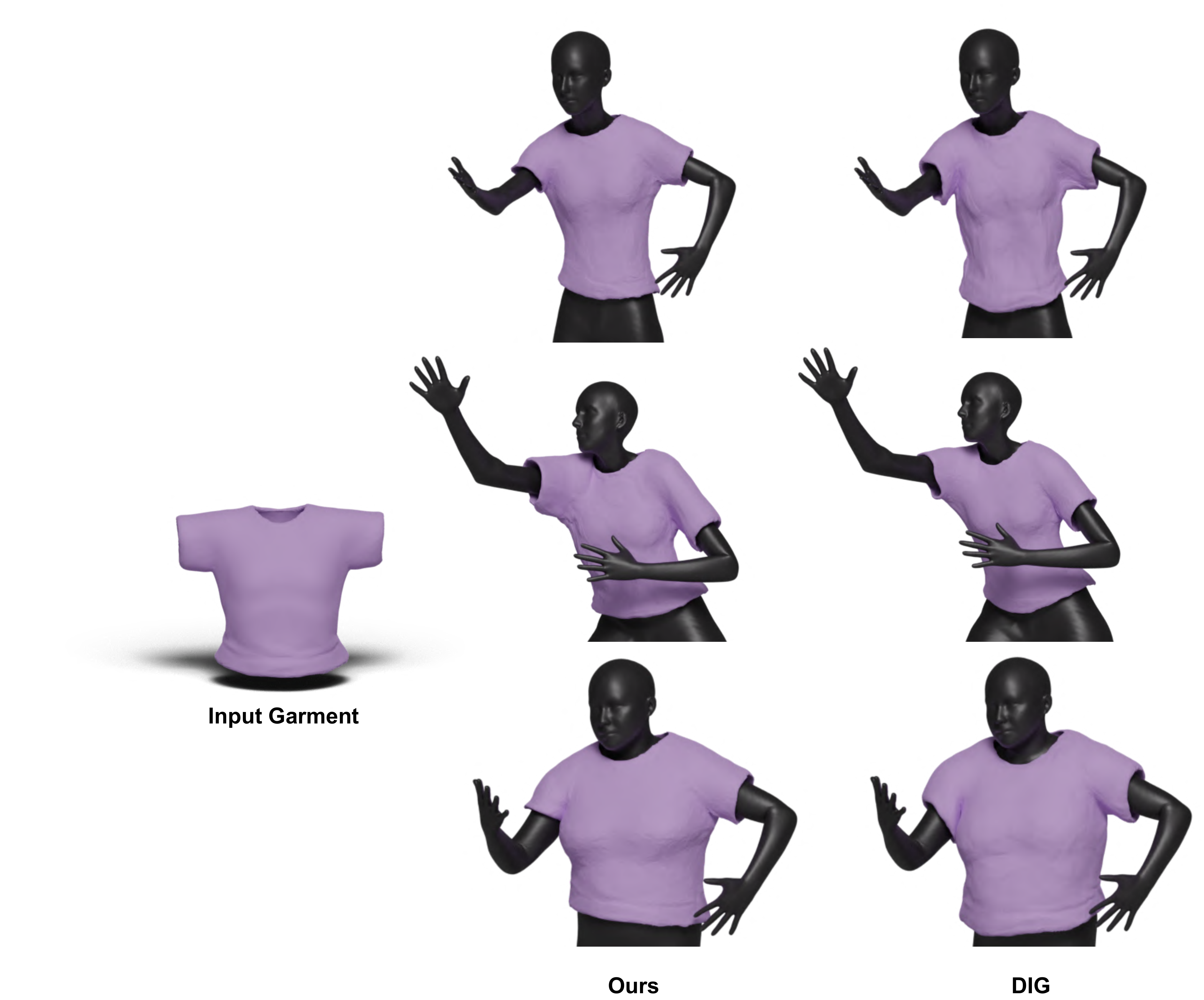}
    % \includesvg[width=0.95\linewidth]{assets/images/dig2.svg  }
    \caption{Qualitative comparison with DIG \cite{dig}.}
    \label{fig:dig}
    \end{figure}
    
    \subsection{Comparison}
    Fig.~\ref{fig:m3d_comp} shows a comparison of M3DVTON\cite{zhao2021m3d} with our framework on random internet images (as mentioned earlier, we use off-the-shelf method \cite{ECON} to extract 3D garments and target human body). It is evident from the figure that since M3DVTON performs retargeting in 2D space, it doesn't produce accurate geometric deformations. Moreover, since it uses a supervised keypoint detection method for initial TPS-based draping, it suffers when the target subject's garment category doesn't match the source garment category. However, our method doesn't suffer from such limitations and can retarget arbitrary garments on arbitrary targets. Fig.~\ref{fig:dig} shows qualitative comparison of our method with DIG\cite{dig}. Our results are qualitatively on par with DIG. However, they cannont drape onto non-parametric bodies.
\begin{figure} 
\centering
\includegraphics[width=0.7\linewidth]{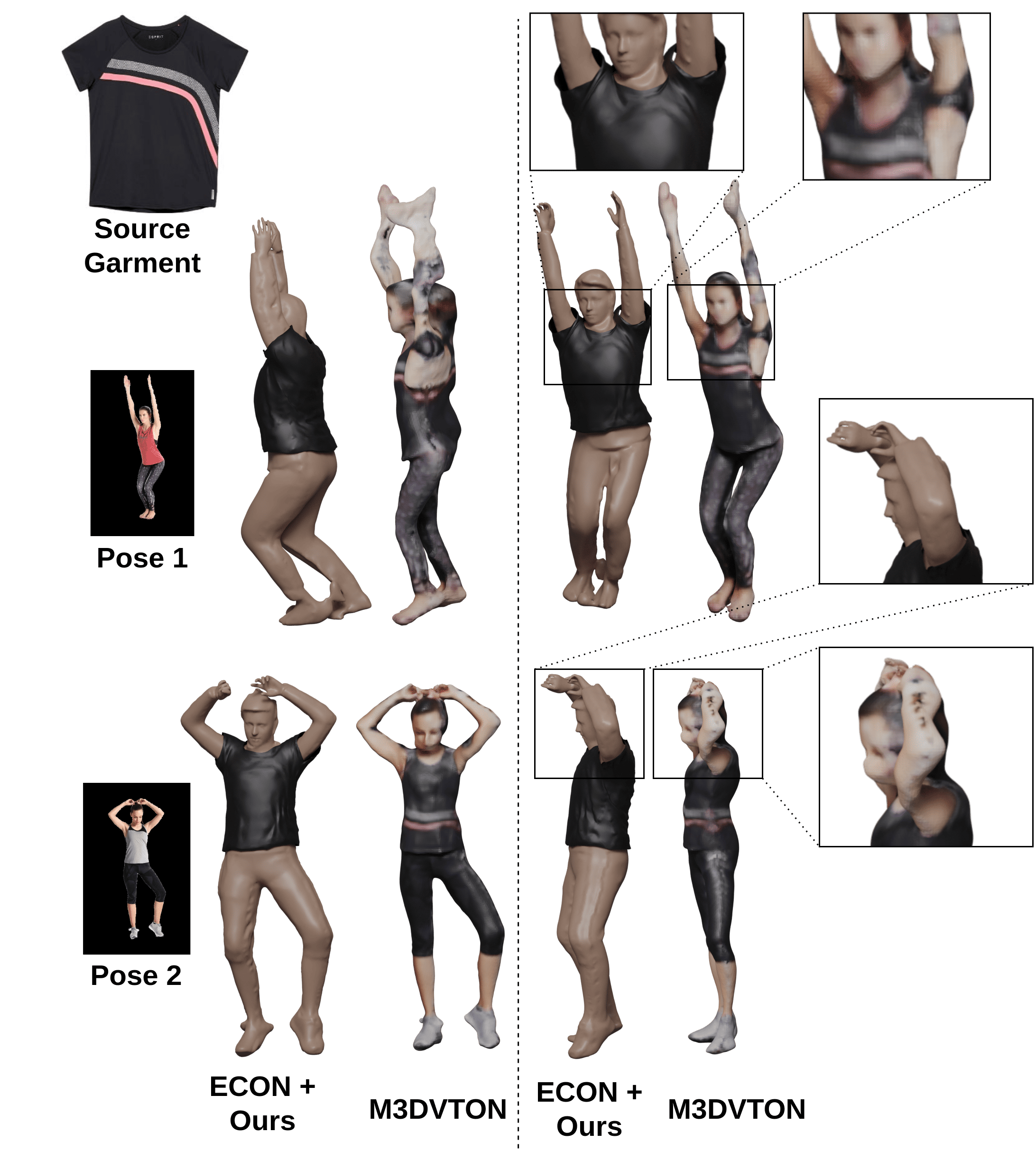}
\caption{Comparison of our method with M3DVTON\cite{M3DVTON} for draping non-parametric garments. M3DVTON introduces false garment geometry (the sleeve of the t-shirt mapped to the sleeveless part of the target geometry) to inaccurate geometries.} 
\label{fig:m3d_comp}
\end{figure}
    
    \begin{figure} 
\centering
\includegraphics[width=\linewidth]{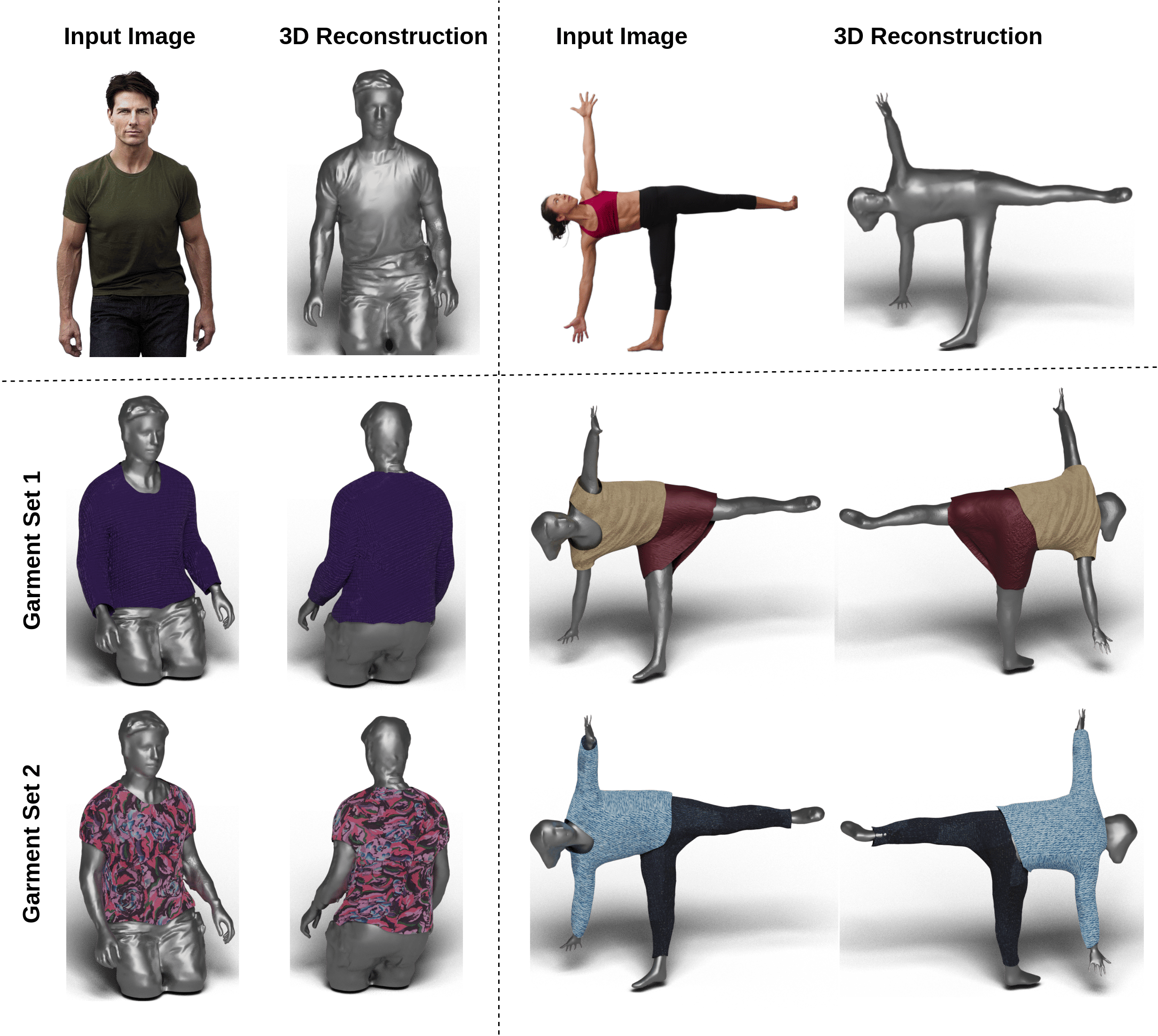}
\caption{Qualitative results of our garment retargeting method on non-parametric avatars reconstructed from internet images.} 
\label{fig:internet}
\end{figure}
    
    % \input{figures/references/8_nsim}

 % \noindent\textbf{Draping Implicit Garments(DIG):} 

    % \noindent\textbf{Neural Cloth Simulation:} 
    %     As discussed in Sec.~\ref{sec:related_work}, simulation-based approaches require a smooth continuous trajectory to repose the garments from one pose to another. When it comes to going from one extreme pose to another (jump trajectory), such methods fail to generate accurate deformations and wrinkles, while also causing severe interpenetrations, as shown in Fig.~\ref{fig:NeuralSim}. Moreover, they offer no provision for draping a garment on an entirely new subject, hence no compliance for 3DVTON solutions. Whereas, our framework can retarget any garment from one extreme pose to another extreme pose, even on an entirely new subject.

    \subsection{Ablation Studies}
   \noindent\textbf{Noise in Correspondence Estimation:}
       Noise in Correspondence Estimation: We analyze the effect of noise in correspondence estimation by introducing noise at different levels. For each correspondence pair$(v_i, x_i)$ we add Gaussian noise to $x_i$ with zero mean and varying standard deviation, i.e. $x_i = x_i + N (0, \sigma)$; $\sigma ={0.001, 0.01, 0.1}$. Please note that for brevity we are writing the 3D noise vector as $N (0, \sigma)$ since $x_i in \mathbb{R}^3$ . We then pass the noisy coarse initialization to the further modules and compute the evaluation metrics (combined for topwear and bottomwear), reported in Table \ref{table:noise_ablation}. As can be seen, our framework is robust enough to handle noise with $\sigma = 0.001, 0.01$, where the evaluation metrics are on par with the noise-free setting. However, with $\sigma = 0.1$, the performance of the method drops.

     \noindent\textbf{Effect of Losses:}
        We also analyze the effect of different loss functions used for self-supervised training for refined retargeting of the garment. \textbf{Please refer to supplementary for additional qualitative and quantitative results}.

   % \subsection{Discussion}

}
\section{Conclusion}
{
    \label{sec:concl}
   We propose a novel,  self-supervised 3D garment retargeting method for non-parametric garments and human body meshes. 
   %Our method is resilient to arbitrary topologies of the garment mesh and variations in human body shapes and poses, due to the robustness in the proposed Isomap embeddings which play a vital role in initial coarse retargeting. 
   We demonstrate high-quality results on both parametric and non-parametric garments/bodies in arbitrary poses and body shapes. 
  % Unlike other methods, we do not impose any constraints on the representation of the input garment. 
   %We believe our method is the first one to support both non-parametric garments and the bodies. 
   Additionally, we also curate a real-world garment dataset to evaluate our method and set a benchmark in non-parametric 3D garment retargeting. %We believe our work will pave the way for the problem of 3D VTON and will prove useful for both industrial and academic purposes.

\newpage
\section{Supplementary Material}

% \tableofcontents
\subsection{Background \& Related Works}
{
    \label{sec:related_work}
We provide a background on different approaches leading towards the problem of 3D virtual tryon while discussing the current landscape and state-of-the-art methods.\\
\\
    \noindent
    \textbf{2D VTON Methods:}  Several 2D VTON methods exist \cite{vton2d1, vton2d2, vton2d3,vton2d4,vton2d5, morelli2023ladi}  which employ deep generative adverserial methods for draping 2D garments over 2D human images. 
    % Specifically, such approaches first aim to segment garments from an input image space, performing an initial thin-plate-spline (TPS) based transformation to roughly warp and align the garment onto  the target image. Then, the transformed garment image is blended with the target person's body image to generate a realistic try-on image, generally using image-to-image translation networks. 
    However, Generative networks tend to produce blurry results and artifacts; even when high-resolution modeling \cite{HR_VTON, HD_VTON} is employed. Moreover, 2D VTON methods have limited ability in terms of adjusting the pose and viewpoint for a more immersive experience. A recently proposed work StylePose \cite{posestyle} has the ability to \emph{repose} the clothed humans to a novel viewpoint in image space leveraging partial 3D priors. However, the work does not allow accurate and view-consistent draping of the 2D garments over a different person altogether, thereby not meeting the basic requirement of a VTON solution. Moreover, to our knowledge, any 2D VTON solution would fail to preserve the accurate view-consistent geometry of the garment after the transformation.\\
    \\
    \noindent
    \textbf{3D VTON Methods:} Clearly, the exploration of 3D space is a more viable option to tackle the aforementioned challenges. 3D-VTON solutions offer the ability to preserve the geometry of the garments and easily allow change of garment and pose properties \& viewpoints. However, there is a significant white space in the area of 3D-VTON research. 3D VTON can be seen as transforming a garment in 3D Euclidean space, in order to align it over (or around) a target 3D human body (SMPL mesh, 3D scan etc.) while avoiding intersections of the garment with the target body. It is highly desirable to model deformations in the garment corresponding to the target body's pose \& shape. Recently, state-of-the-art works like \cite{M3DVTON} claim to propose the first 3D VTON solution by extending the 2D TPS-driven generative pipeline to reconstruct the 3D geometry, finally blending on a try-on image, with a representation similar to that of Moulding-Human \cite{mouldinghumans}. Although this allows viewing the draped garment on the target body from arbitrary viewpoints, the draping is still performed in image space using GANs and hence suffers from limitations such as blurry artifacts and false geometrical deformations. Additionally, since the method starts from the image of a garment, extending it to a real-world scan of a 3D garment is not trivial.\\
    \\
    \noindent
    \textbf{Neural Garment Simulation:} Researchers have proposed learning-based garment simulation methods for increasing efficiency and speed for modeling garment dynamics, as the classical physics-based simulation \cite{nealen2006pbs} is computationally expensive and slow.  At first glance, it looks like such methods have the capability to perform 3D garment retargeting. However, it is important to note that \textit{\textbf{garment retargeting is a different problem than garment simulation}}. In simulation, the goal is to realistically deform the garment gradually as the underlying body dynamically changes the pose over an animated sequence. It assumes a complete trajectory of the underlying body going from an initial pose to a final pose. On the other hand, garment retargeting deals with the transfer of a garment from one pose to another, even on a different subject altogether. State-of-the-art neural garment simulation methods \cite{SNUG, neural_cloth_sim, HOOD} aim at self-supervised draping of a parametric garment on top of a given parametric body sequence evolving over that time. The self-supervision comes from the physics-inspired constraints during the loss minimization.  While the simulation-based approaches provide an accurate detailing of deformation and wrinkles, they often rely on the previous frames to obtain simulation-specific parameters, e.g. velocity and acceleration information. If we directly try to retarget (or simulate in this case) the garment from one pose to another arbitrary pose, such an approach suffers drastically due to not enough motion information between the source garment pose and the target body pose. Additionally, they do not support changing the shape/subject in between the simulation. In contrast, 3D garment retargeting aims at transforming the vertices of a garment mesh to drape it over a target body of arbitrary pose/shape directly in one shot without requiring underlying body pose sequences.
    % \begin{figure}[!h]
    % \centering
    %     \includegraphics[trim={8cm 0 0 8cm},clip, width=1.1\linewidth, trim]{figures/HOOD_gap.jpg}
    %     \caption{Failure of SOTA neural garment simulation-based methods to perform retargeting of the 3D garment from one arbitrary pose to the other when intermediate poses are unavailable.}
    %     \label{fig:hood_gap}
    % \end{figure}
    \\
    \noindent
    \textbf{Physics Inspired Garment Draping:} Some of the recent deep learning-based efforts like \cite{deeppsd} have made progress in this direction utilizing supervised training strategies learning the skinning weights of the parametric garment for draping it onto a parametric human body. They consider SMPL~\cite{SMPL:2015} as the parametric body model, and garments are also derived from the SMPL body mesh~\cite{patel20tailornet, dig, drapenet}. All the aforementioned methods \textit{\textbf{don't perform retargeting from scratch}}, i.e. they need a 3D garment already \textbf{perfectly fitted} on top of a parametric body in rest pose (T-pose or A-pose), or alternatively a latent encoding of the garment. These methods are trained in a self-supervised fashion using physics-based constraints to predict the deformation in the canonical/latent garment according to the shape and pose of the underlying parametric body. Moreover, in order to train such methods, a large number of change parametric garments in canonical pose and shape are required to obtain the latent representation. Additionally, they don't support non-parametric garments, e.g a garment and body extracted from a real scan or reconstructed from an image (using \cite{reef, xcloth,deep_parametric_singleview}) are non-parametric in nature and the aforementioned approaches cannot handle them.\\
    \\
    % On the other hand, several deep learning methods \cite{SNUG, HOOD} have been proposed, which learn to simulate a 3D garment mesh onto a 3D body mesh, \cite{SNUG} and \cite{neural_cloth_sim} rely on self-supervised physics-based losses in order to model the dynamics of a garment due to changes in the underlying body pose. More specifically, these methods learn to transform the 3D garment mesh over a gradually animating 3D body (a smooth pose-change trajectory). Such a formulation, replicating cloth simulation, will not be able to deal with sudden changes in the pose and shape.
    %{\color{blue} With increasing availability of 3D human meshes, the 3D instance try-on is the most desired over the simulated stack as discussed in Sec.\ref{sec:intro}. 
    % Methods like TailorNet\cite{patel20tailornet} have made promising progress towards 3D instance try-ons and follow-up works \cite{smplicit, drapenet, dig} have extended incorporating a corpus of garments as a latent representation. However, all of them rely on synthetic or parametric garments rather than dealing with real-world scanned garments.
    % Another class of methods \cite{DIG, DrapeNet} require skinnning wirghts requiring specific joints
    %
    \noindent
    While it is true that extending these works to real-world garments is challenging, validation of the leveraged technique is also a significant challenge. As most commonly available multi-pose clothed-human datasets either provide synthetic and parametric clothing\cite{bertiche2020cloth3d, 3dpeople}  or lack garment-specific shape variation\cite{amass, 3DHumansDataset}.
}

    \subsection{Implementation Details}
    {
        \label{sec:impl_details}
        \subsubsection{SMPL Registration:}
        {
            \label{sub_sec:smpl_registration}
            In order to establish the dense correspondences for coarse retargeting of the mesh, we first estimate the pose \& shape of the underlying body in both meshes (the \emph{garment} as well as the \emph{target} body). 
            % The pose of the SMPL is defined by vector $\theta = [\theta_0, \theta_1,..., \theta_{23}]$ where $\theta_i \in \mathbb{R}^3$ denotes the axis-angle representation of the relative rotation of part $i$ with respect to its parent in the kinematic tree. The shape parameters are defined by the vector $\beta \in \mathbb{R}^{10} $, which governs the prominent shape changes and is estimated by taking PCA over a large set of 3D human scans.
            If the garment or the target body is already present in canonical pose and shape, then the SMPL parameters can be directly picked from the canonicalized SMPL. In the absence of canonicalized meshes (garments or target bodies), we employ a similar SMPL fitting strategy as proposed by \textsc{PaMir}\cite{PaMir} for obtaining SMPL body parameters. The pipeline of \textsc{PaMir} extends the SMPL fitting methodology of \cite{smplify-x}, exploiting multi-view consistency. The resultants are registered SMPL bodies for both the garment and target-body meshes. \emph{It is to be noted that, despite massive efforts to employ multi-view consistency, the registration pipeline is far from accurate}. Our framework is robust enough to handle noise in pose \& shape parameters. Finally, the estimated pose \& shape parameters are used to generate SMPL mesh $\mathcal{M}$, consisting of $6,890$ vertices and $13,776$ faces. This step is important for estimating isomap embeddings for each vertex of the garment using k-nearest-neighbor extrapolation of SMPL vertices.
        }
        
        \subsubsection{Refined Retargeting Module}
        {
            \label{sub_sec:refined_retargeting}
            The coarse retargeted mesh obtained using dense correspondence between garment and target body is refined using a self-supervised \emph{Refined Retargeting Module}. It is composed of two PointNet encoders $PointNet_{\mathcal{G}}$ and $PointNet_{\mathcal{T}}$ for encoding both input garment and target body, respectively and an MLP decoder. The PointNet encoder consists of 5 ResNet blocks with skip connections between each block. Each ResNet block is an FC (fully connected) layer with ReLu activations. Each encoder outputs a latent code of $128$-dimension. These encodings, along with the coarsely initialized garment vertices, \textit{k}-neighbors of target mesh, and the iso-embedding of the input garment are fed to the MLP decoder. The MLP is constituted of six hidden layers with 512 neurons, each activated by LeakyReLu functions. The last layer of MLP is a Tanh. 
            
            Apart from feeding PointNet features of the garment and body as input, we also condition every layer of the MLP with PointNet features similar to ADAIN\cite{adain}. The MLP outputs a $\Delta x$ value, which is added to the \emph{course-retargeted} mesh to obtain \emph{refined-retargeted} mesh.
            
        }
        
        % \subsection{Wrinkle Generation}
        % {
        %     \label{sub_sec:wrinkle_generation}
        %     As an optional post-processing step, we learn to induce plausible wrinkles on the garment mesh, conditioned on the target body pose and shape. The smooth input garment is encoded using DiffusionNet \cite{diffusionNet} which is effective in learning high-frequency details. The DiffusionNet consists of 4 blocks, with a channel width of 128 and 128 eigenbasis vectors for spectral acceleration. We use ReLU activations at intermediate layers and softmax for the final layer. We also use PointNet encoders, similar to one used in \emph{Refine Retargeting Module} for encoding target vertices. We use a decoder MLP with six hidden layers of 512 neurons each. The MLP is activated by LeakyReLu activation in the internal layers and with a Tanh final layer. The MLP takes the DiffusionNet and PointNet encodings as input and outputs a $\delta$ value per vertex which is added to the input-smooth garment to obtain plausible wrinkles. Refer \ref{fig:wrinkle_generation} 
        %     \input{figures/suppl_figures/wrinkle_generation_pipeline}
            
        % }

    }

    \subsection{Extended Qualitative Results}
    {
        \label{sec:extended_qual_res}
        \noindent
        In this section, we discuss extended qualitative results in various data settings. Please refer to the supplementary video for 360-degree renderings of the results.
        \subsubsection{CLOTH3D Garments on SMPLs of AMAAS Data}
        {
            \label{sub_sec:cloth3d_smpl}

            \begin{figure*} 
\centering
\includegraphics[width=\linewidth]{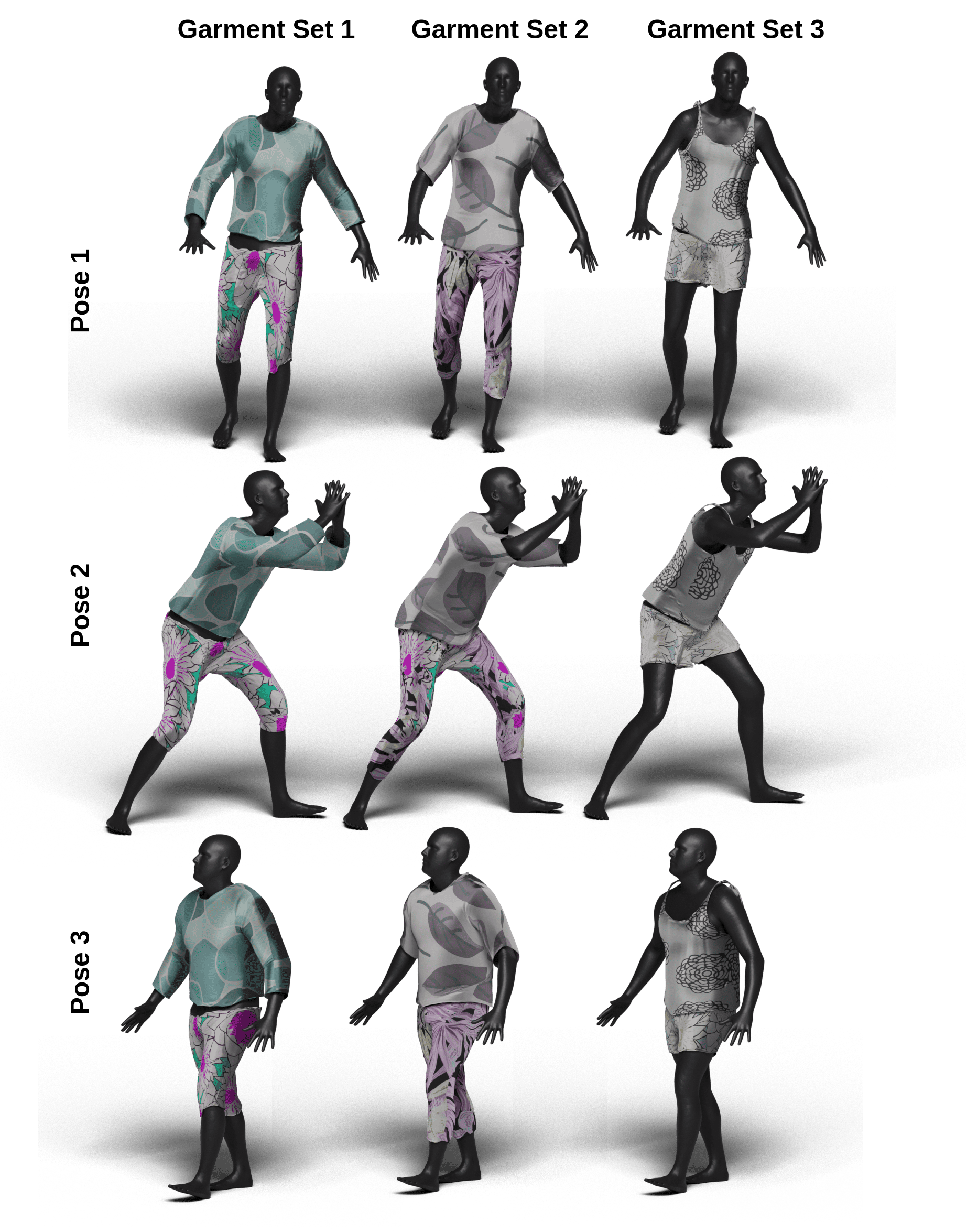}
\caption{Cloth3D garments draped on smpl samples from AMAAS dataset} 
\label{fig:cloth3d_amaas_new}
\end{figure*}
    
            In Fig.\ref{fig:cloth3d_amaas_new}, we show qualitative results of our method on CLOTH3D data, which is draped onto three distinctive and challenging SMPL poses obtained from AMAAS\cite{amass} dataset. Do note that we also demonstrate our results of bottom wear. 
        }
        % \subsection{CLOTH3D Garments on Real Scans}
        % {
        %     \label{sub_sec:cloth3d_real}
        %     % \input{figures/suppl_figures/cloth3d_real_scans}
        %     % \input{figures/references/3dhuman_thuman}
        %     We also show qualitative results of our method on real scans of the THumans2.0 dataset in Fig.\ref{fig:cloth3d_real_scans}. Despite changes in the topology of the body, our method is able to generalize well to the scanned human body, which further validates our method. Please note that as we are dealing in mesh space, our method can handle arbitrary representations of the body, be it scanned or SMPL. 
        % }
        \subsubsection{DressMeUp Garments on SMPLs}
        {
            \label{sub_sec:DMU_smpl}
            \input{figures/suppl_figures/dmu_on_smpl}
            In Fig.\ref{fig:dmu_smpl}, we show our real-world scan being draped onto SMPLs of AMAAS data.
            
        }
        \subsubsection{DressMeUp Garments on Real Scans}
        {
            \label{sub_sec:DMU_real}
            \begin{figure*}
    \begin{tikzpicture}[scale=0.9]
    \node (A) at (-.5,0) {\includegraphics[width=1.05\textwidth]{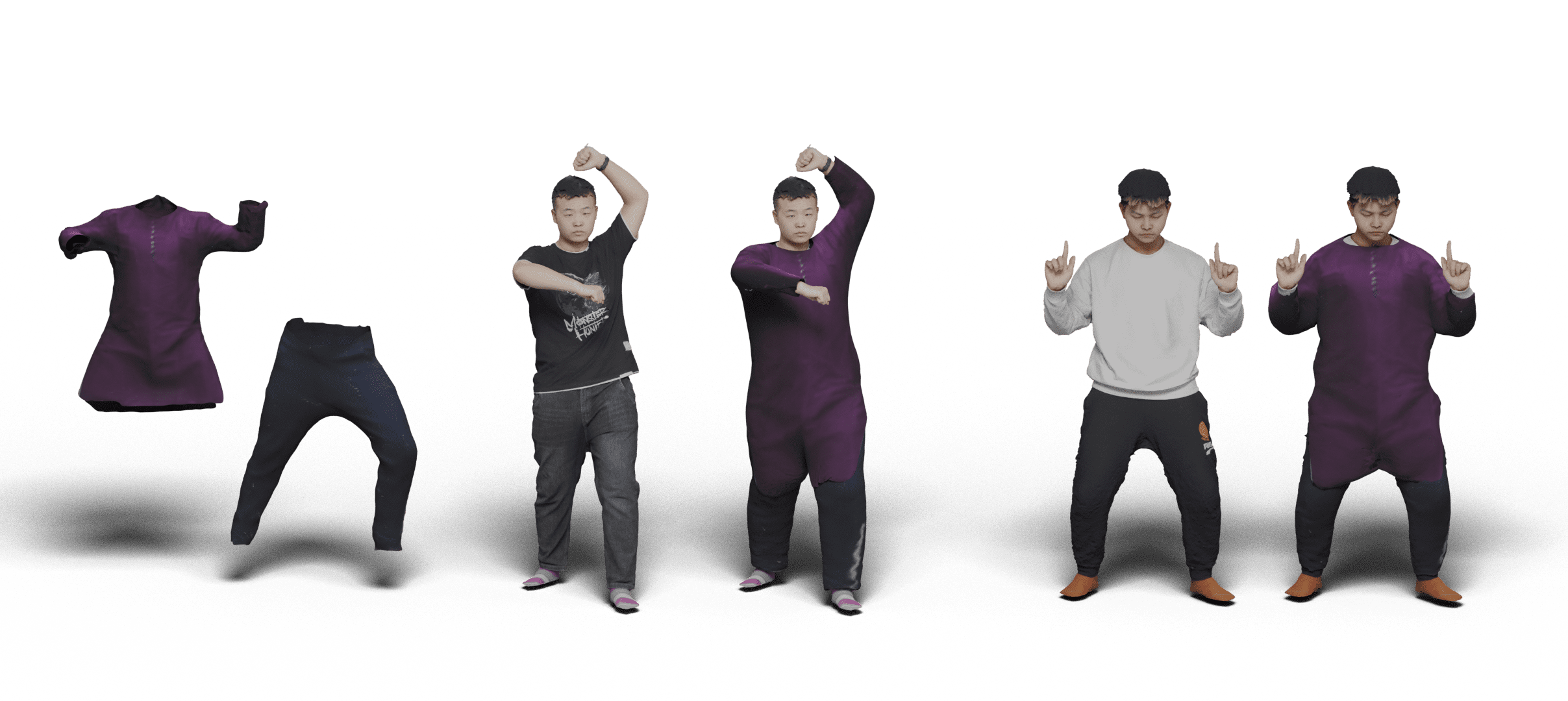}};
    \node (B) at (0,-7) {\includegraphics[width=\textwidth]{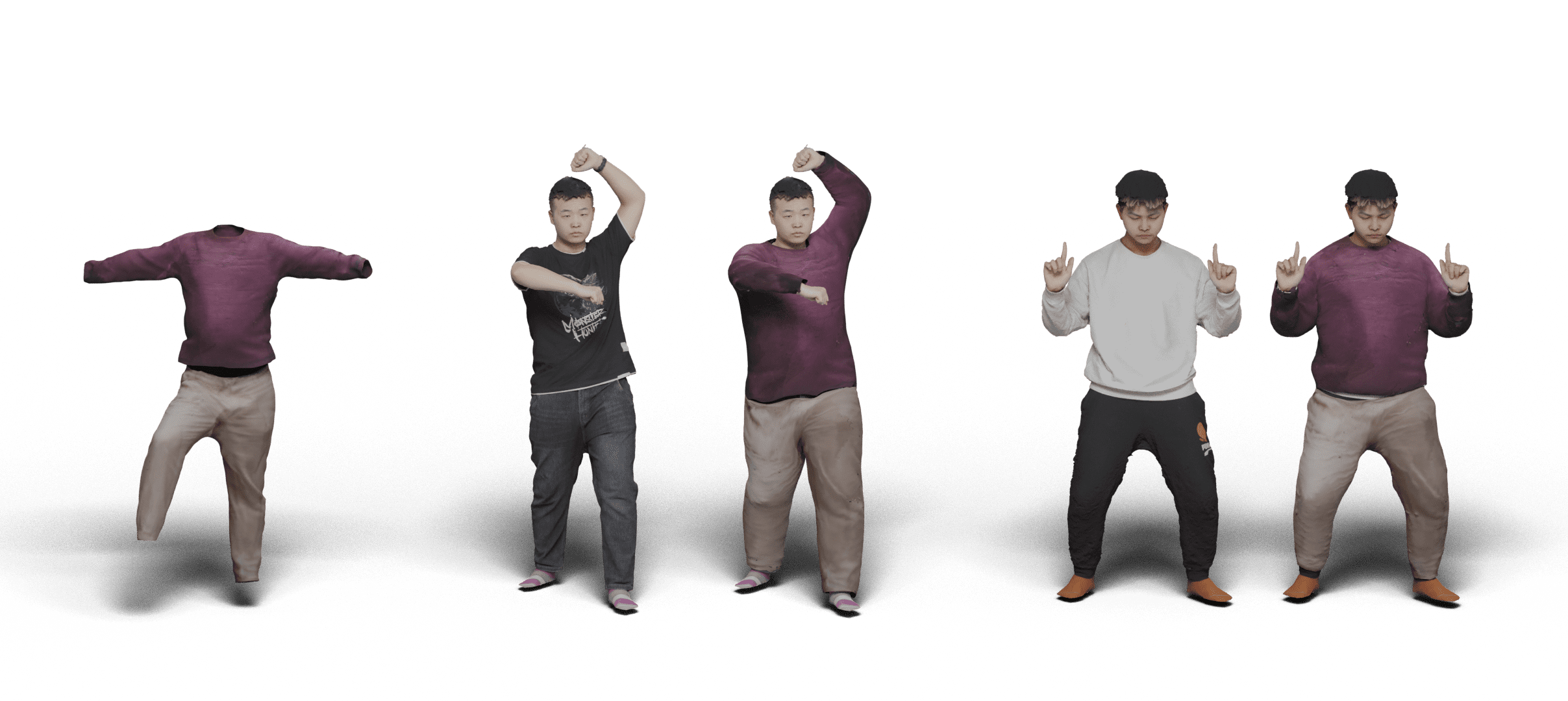}};
    \node[below, black] at (-7, -10.5) {\small (a)};
    \node[below,black] at (-2.5, -10.5) {\small (b)};
    \node[below,black] at (-0, -10.5) {\small (c)};
    \node[below,black] at (4.5, -10.5) {\small (d)};
    \node[below,black] at (7.5, -10.5) {\small (e)};
    \end{tikzpicture}
    \caption{The figure shows different real scanned garments of our \emph{Dress Me Up} dataset draped onto real-scans of T-humans2.0 human body scans, (a) shows the \emph{Dress Me Up}'s real-garments and columns (b) and (d) show scanned humans of Thumans2.0, we employ our proposed framework to drape these real garments to arbitrary real body scans of Thumans2.0 dataset as visualized in columns (c) and (e).}
    \label{fig:dmu_real_scans}
\end{figure*}
            We show the results of DressMeUp garments draped on real scans of THuman2.0 dataset. Refer \ref{fig:dmu_real_scans}. Our method produces plausible retargeting of data scans.
        }
        % \subsection{Results on Internet Images}
        % {
        %     \label{sub_sec:internet_images}
        %     \input{figures/suppl_figures/internet_images}
        %     We show additional results on internet images. We reconstruct 3D human models from internet images using existing methods and also extract garments from these reconstructed scans. Visualization of arbitrary garments obtained from internet images draped on to reconstructed human models is shown.
        % }
        % \subsection{Extended Comparison with M3DVTON}
        % {
        %     \label{sub_sec:m3dvton_ext}
        %     We show additional results in comparison with the M3DVTON method.

        %     \input{figures/suppl_figures/m3dvton}
    
        % }
        % \subsection{Extended Comparison with DIG}
        % {
        %     \label{sub_sec:dig_ext}
        % }
    }

    \subsection{Description of Evaluation Metrics}
    {
        \label{sec:eval_metrics}
        \noindent
        Given a 3D garment mesh $\mathcal{G}$ to be retargeted and the corresponding GT garment mesh $\mathcal{G}_{GT}$ (where $v_i\in vertices(\mathcal{G})$ and $\hat{v}_i \in vertices(\mathcal{G}_{GT}))$, we use the following standard metrics for evaluation:

        \noindent\textbf{Euclidean Distance(ED):} We compute ED as the average Euclidean distance between the corresponding vertices of input and final retargeted garment mesh, i.e.
        \begin{equation}
            ED = \frac{1}{n} \sum_{i=1}^{n} \left\Vert v_i - \hat{v}_i \right\Vert
        \end{equation}
        Lower values for ED are desired for better output.
            
       \noindent\textbf{Normal Consistency(NC):} We compute NC as the average cosine similarity between the corresponding vertex normals of input and final retargeted garment mesh, i.e.
        \begin{equation}
            NC = \frac{1}{n} \sum_{i=1}^{n} n_i \cdot \hat{n}_i
        \end{equation}
        Values close to $1$ are desirable for NC.
            
        \noindent\textbf{Interpenetration Ratio(IR):} It is computed as the ratio of the area of garment faces
        inside the body to the overall area of the garment faces; hence lower values are desired to ensure the least amount of penetration of the garment mesh with the target body mesh.
    
            \textbf{Chamfer Distance (CD):} Given two sets of points $S_1$ and $S_2$, Chamfer distance measures the discrepancy between them as follows:
    
            \begin{equation}
                            \begin{aligned}
                                CD &= \sum_{x \epsilon S_1} min_{y \epsilon S_2} \Vert x-y \Vert _2^2 \\
                                &+ \sum_{y \epsilon S_2} min_{x \epsilon S_1} \Vert x-y \Vert _2^2\\
                            \end{aligned}
                        \label{eq:CD}
            \end{equation}
            In our case, $S_1 = vertices(\mathcal{G})$ and $S_1 = vertices(\mathcal{G}_{GT})$.
            
        \noindent\textbf{Point-to-Surface (P2S) Distance:} P2S measures the average L2 distance between each vertex of the garment mesh and the nearest point to it on the target body surface. 
    }

    \subsection{Extended Ablation Study}
    {
        \label{sec:ablations}
        \noindent
        In this section, we discuss the ablation of self-supervised losses of the refinement module.

        % \subsection{Ablation on Self-Supervised Losses}
        % {
        %     \label{sub_sec:ablation_self_supervised}
            We provide an ablative study of the effect of each loss in the Refined Retargeting module and report the relevant metrics in Table.\ref{table:loss_ablation}.  
        % }
        % \subsection{Ablation on Wrinkle Generation Module}
        % {
        %     \label{sub_sec:wrinkle}
        %     \input{tables/6_wrinkle_ablation}
        %     We provide a quantitative evaluation of the performance of our wrinkle generation network on two datasets 3DHumans \cite{3DHumansDataset} and THuman2.0 \cite{thuman}. We divide both the datasets into train and test splits and report the P2S distance, Euclidean distance and Normal Consistency losses. Refer to Table.\ref{table:wrinkle_tab}
        % }
    }

    \subsection{Discussion}
    {
        \label{sec:discussion}
        \subsubsection{Description of DressMeUp Dataset}
        {
            \label{sub_sec:dataset_desc}
            \input{figures/suppl_figures/dress_me_up}
            We provide our own textured garment dataset, curated using Kinect cameras. The dataset consists of 50 different garments, with 44 unique garments worn by 15 individuals. Each garment is provided in 5 different poses on the same person, resulting in a total of 250 garment meshes. The garments category include full and half-sleeved Tshirts, Trousers, half-pants, kurta, dress, open shirt etc.
        }
         \begin{table}
            \centering
            \begin{tabular}{|l |c |}
            \hline
            {\textbf{Representation}} & {\textbf{$\mathcal{R}_{score}\downarrow$}}\\
            \hline
            {BodyMap\cite{bodymap}} & {0.955}\\
            {16-dim. Isomap Embeddings} & {0.491} \\
            {32-dim. Isomap Embeddings} & {0.473} \\
            {64-dim. Isomap Embeddings} & {0.437} \\
            {128-dim. Isomap Embeddings} & {0.426} \\
            {256-dim. Isomap Embeddings} & {0.424} \\
            \hline
            \end{tabular}
            \caption{Analysis of choice of representations for correspondence estimation. $\mathcal{R}_{score}$ takes values between 0 \& 1, where lower values are preferred.}
            \label{table:iso_analy}
        \end{table}
        
        \begin{figure}
            \centering
            \includegraphics[width=\linewidth]{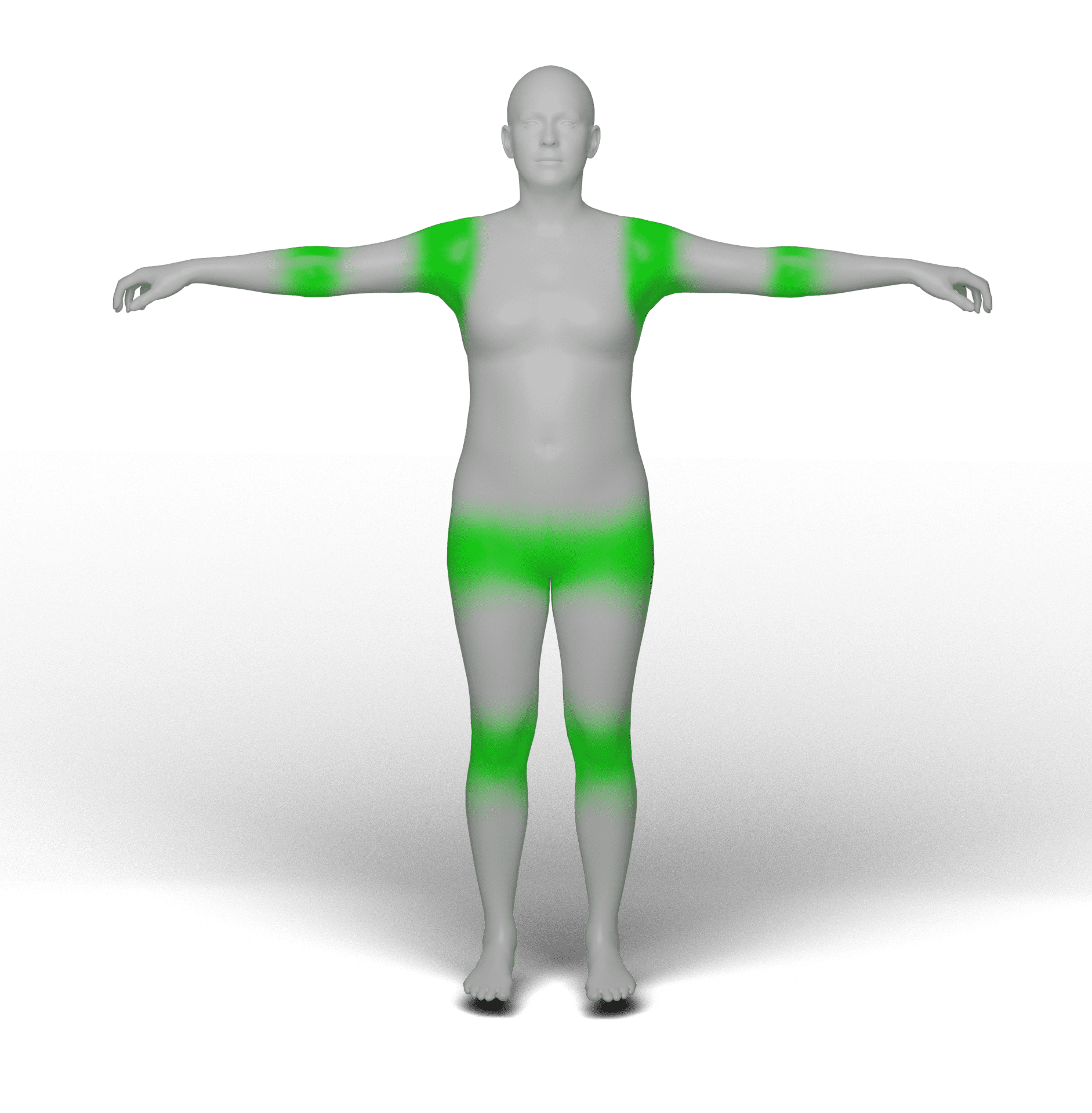}
            \caption{Joint Masks}
            \label{fig:joint_mask}
        \end{figure}
        
        \subsubsection{Analysis of Isomap Embeddings}
        {
        We propose a novel strategy that allows establishing correspondences between different human scans, garments, or anything that resembles human body structure. SMPL being a parametric human body model, acts as a reasonable medium to establish correspondences across different body shapes, poses, and appearances. As explained in the main draft, once both the garment and the target body (parametric or non-parametric) are registered with SMPL, where the target body can be an SMPL mesh itself, we compute 128-dimensional isomap embeddings for each vertex of the garment and target body. Then, dense correspondences can be established between the two by matching similar 128-dimensional extrapolated features.

        We arrive at this choice of feature modeling after carefully studying existing representations for dense correspondence matching for humans. This problem is specifically tough as humans are deformable objects and tend to undergo non-rigid motion. Continuous Surface Embeddings (CSE)\cite{ContinuousSE} propose a learnable image-based representation of dense correspondences and a model which predicts, for each pixel in a 2D image, an embedding vector of the corresponding vertex in the object mesh, therefore establishing dense correspondences between image pixels and 3D object geometry. The authors show remarkable results in matching correspondences across RGB human images via 16-dimensional representation vectors. Recently, BodyMap\cite{bodymap} proposed to extend this approach by extrapolating the CSE embeddings of SMPLs registered with high-quality human scans in UV space. We started with BodyMap representation but later found it to produce a lot of false matching, and we decided to analyze the behavior quantitatively.

        The representation for correspondence estimation should be rich and varied enough to avoid repetitions in the feature space when extrapolated, otherwise, different body parts would map nearby in the embedding space. More specifically, geodesically far-apart vertices should map far apart in the embedding space and vice-versa. Based on this ideation, we design an evaluation metric,\textbf{\textit{ Richness Score}}($\mathcal{R}_{score}$) for each vertex $v_i$ of SMPL mesh, which is calculated as follows:
        
        \begin{equation}
            \mathcal{R}_{score_i} = (\mathcal{R}_{near_i} + \mathcal{R}_{far_i}) / 2
        \end{equation}
        \begin{equation}
            \mathcal{R}_{near_i} = \frac{1}{k^2} \sum_{i=1}^k min(| \mathcal{N}_{geo}^{rank} - \mathcal{N}_{emb}^{rank} |, k)
        \end{equation}
        \begin{equation}
            \mathcal{R}_{far_i} = \frac{1}{k^2} \sum_{i=1}^k min(| \mathcal{F}_{geo}^{rank} - \mathcal{F}_{emb}^{rank} |, k)
        \end{equation}
        
         where, $\mathcal{N}_{geo}^{rank}$ \& $\mathcal{N}_{emb}^{rank}$ denotes the ranks of k-nearest neighbors of $v_i$ in both geodesic and embedding space, and similarly, $\mathcal{F}_{geo}^{rank}$ \& $\mathcal{F}_{emb}^{rank}$ denotes the ranks of k-farthest neighbors of $v_i$ in both geodesic and embedding space. Thus, $\mathcal{R}_score$ penalizes if the rank of neighbors (k-nearest and k-farthest) in geodesic and embedding space doesn't match. We report the values in Table.\ref{table:iso_analy}, where it can be seen that extrapolating isoembedding values in Euclidean space has a better effect than BodyMap\cite{bodymap}. The remaining values show that high dimensionality is preferred. However, empirically, values are saturated once a significant dimensionality is reached.
        
            \label{sub_sec:isomap_embed}
        }
        \subsubsection{Applications of the Proposed Framework}
        {
            \label{sub_sec:applications}
            \begin{itemize}
                \item \textbf{3D VTON for Arbitrary Garments}
                Our proposed framework can be seen as a potential solution for 3D VTON problem. As evident from our qualitative results, the proposed framework can generalize well to unseen real and non-parametric garments, and retarget them to arbitrary posed and shaped human scans.
                \item \textbf{Size-fitting Solutions}
                It is important to note that although we aim to preserve the overall structure of the garment to be retargeted, the final garment could scale accordingly to the target body. This is actually preferred as different people wear different sizes (M, L, XL, XXL) of the garments of the same style. Our framework can drape garments to arbitrary sizes (need not be discreet) which is a unique contribution to the size-fitting solution.
                \item \textbf{Layered Clothing:} As can be seen from our qualitative results on real scan, we can easily retarget garments on top of humans already wearing garments, thereby enabling layered clothing, which is an extremely challenging task.
                \item \textbf{Generating Ground Truth Data for 2D VTON Methods}
                Since, we can retarget the 3D garment into different poses and even on different subjects, and eventually can render them consistently in the form of 2D images, our framework can easily be used for generating photorealistic high-quality 2D VTON datasets from a limited number of 3D data samples. This is another highly useful application of our framework, and we intend to use it to develop and release such large-scale datasets in the public domain to accelerate the 2D VTON research as well.
            \end{itemize}
        }
        \subsubsection{Limitations \& Future Work}
        {
            \label{sub_sec:limitations}
            We proposed a method for self-supervised 3D garment retargeting and a first-of-its kind 3D VTON dataset for evaluating our framework. We showed that our novel framework leverages the isomap via SMPL to establish dense correspondences and initial coarse retargeting, which is then used as a prior for training a self-supervised learning technique for refining the retargeting. Being the first method for retargeting (not just neural rendering) the 3D non-parametric garment mesh from real-world distribution, we qualitatively show superior performance to similar State-of-the-Art methods.

            Although we can retarget 3D garments on top of arbitrary human scans, currently there is no provision to remove the underlying garment the subject is already wearing. However, this is an extremely complex task as it might require reconstructing the underlying human body (for e.g. if a half t-shirt is to be draped over a subject wearing full t-shirt, removing full t-shirt requires reconstructing the arms of the subject). Though, we can easily handle noisy SMPL registration, small penetration noise can be noticed when the geometry of the input garment is bad, especially when the garment is reconstructed from RGB image using off-the-shelf networks (e.g. \cite{reef}). Finally, we aim to model extremely loose and free-flowing garments, such as long gowns, \textit{sarees}, etc. We hope our method paves the way for handling the aforementioned problems we would like to tackle in the future.
            
        }
        \subsubsection{Supplementary Video}
        {
            \label{sub_sec:suppl_video}
            Please refer to the supplementary video for a better understanding of the approach and qualitative results, where we provide 360-degree visualizations of the figures.
        }
    }

     \begin{table}
    \centering
    \begin{tabular}{|l |c c| c c|}
    \hline
    \multirow{2}{*}{Loss type} & {P2S$\downarrow$} & \multicolumn{1}{c|}{ED$\downarrow$} & {NC$\uparrow$} & {IR\%$\downarrow$}\\
                           & \multicolumn{2}{c|}{x $10^{-3}$} & &\\
    \hline
    {$\mathcal{L}_{corres}$ only} & {7.406} & {9.593}  & {0.935} & {0.0217}\\
    
    {$\mathcal{L}_{length}$} & {9.614} & {11.352}  & {0.932} & {0.058}\\
    
    {$\mathcal{L}_{bend}$ only} & { 10.245} & {11.923}  & {0.928} & {0.104}\\

    {Without $\mathcal{L}_{corres}$} & { 12.125} & {13.445}  & {0.929} & {0.135}\\
                      
    {Without $\mathcal{L}_{length}$} & { 10.560} & {11.940}  & {0.933} & {0.022}\\

    {Without $\mathcal{L}_{bend}$} & { 7.406} & {9.593}  & {0.935} & {0.021}\\
    
    {Without Joint Mask} & {10.560} & {11.941}  & {0.933} & {0.022}\\

    \hline
    \end{tabular}
    \caption{Quantitave evaluation of Wrinkle Generation Network}
    \label{table:loss_ablation}
\end{table}

     \begin{table}
    \centering
    \begin{tabular}{|l |c c| c c|}
    \hline
    \multirow{2}{*}{Loss type} & {P2S$\downarrow$} & \multicolumn{1}{c|}{ED$\downarrow$} & {NC$\uparrow$} & {IR\%$\downarrow$}\\
                           & \multicolumn{2}{c|}{x $10^{-3}$} & &\\
    \hline
    
    {10 garments} & {6.901} & {9.353}  & {0.951} & {0.009}\\

    {50 garments} & {7.370} & {9.511}  & {0.934} & {0.008}\\
    
    \hline
    \end{tabular}
    \caption{Evaluation of network trained with 10 and 50 Cloth3D garments and evaluated on test samples.}
    \label{table:num_gar}
\end{table}

    % {\small
    %     % \bibliographystyle{ieee_fullname}
    %     % \bibliography{main}
        
    %     \bibliographystyle{ieeenat_fullname}
    %     \bibliography{main}
    % }

% \clearpage
{
    \small
    \bibliographystyle{ieeenat_fullname}
    \bibliography{main}
}

\end{document}